\documentclass[10pt,journal,compsoc]{IEEEtran}
\ifCLASSOPTIONcompsoc
  \usepackage[nocompress]{cite}
\else
  \usepackage{cite}
\fi
\ifCLASSINFOpdf
\else
\fi
\usepackage{amsmath,amsfonts}
\usepackage{algorithmic}
\usepackage{algorithm}
\usepackage{array}
\usepackage{textcomp}
\usepackage{stfloats}
\usepackage{url}
\usepackage{verbatim}
\usepackage{graphicx,subcaption}
\usepackage{cite}
\usepackage{tikz}
\usepackage{booktabs}
\usepackage{multirow}
\usepackage{amsmath}
\DeclareMathOperator*{\argmax}{arg\,max}

\makeatletter
\newcommand*{\rom}[1]{\expandafter\@slowromancap\romannumeral #1@}
\makeatother

\begin{document}
%
\title{Simple Primitives with Feasibility- and Contextuality-Dependence for Open-World Compositional Zero-shot Learning}
%
%
%
%

\author{Zhe~Liu,
        Yun~Li,
        Lina~Yao~\IEEEmembership{Senior~Member,~IEEE},
        Xiaojun~Chang~\IEEEmembership{Senior~Member,~IEEE},
        Wei~Fang~\IEEEmembership{Member,~IEEE},
        Xiaojun Wu~\IEEEmembership{Member,~IEEE},
        and Yi Yang~\IEEEmembership{Senior~Member,~IEEE}
\thanks{Z. Liu, W. Fang, and X. Wu are with Jiangsu Provincial Engineering Laboratory of Pattern Recognition and Computational Intelligence, Jiangnan University. E-mail: zheliu912@gmail.com; fangwei@jiangnan.edu.cn; wu$\_$xiaojun@jiangnan.edu.cn. }
\thanks{Y. Li and L. Yao are with the School of Computer Science and Engineering, University of New South Wales. E-mail: yun.li5@unsw.edu.au; lina.yao@unsw.edu.au. }
\thanks{X. Chang is with the Australian Artificial Intelligence Institute, University of Technology Sydney. Email: xiaojun.chang@uts.edu.au. }
\thanks{Y. Yang is with School of Computer Science and Technology, Zhejiang University. Email: yangyics@zju.edu.cn. }
}

%
%

\markboth{}%
{}
%



\IEEEtitleabstractindextext{%
\begin{abstract}
The task of Compositional Zero-Shot Learning (CZSL) is to recognize images of novel state-object compositions that are absent during the training stage. Previous methods of learning compositional embedding have shown effectiveness in closed-world CZSL. However, in Open-World CZSL (OW-CZSL), their performance tends to degrade significantly due to the large cardinality of possible compositions. Some recent works separately predict simple primitives (i.e., states and objects) to reduce cardinality. However, they consider simple primitives as independent probability distributions, ignoring the heavy dependence between states, objects, and compositions. In this paper, we model the dependence of compositions via feasibility and contextuality. 
Feasibility-dependence refers to the unequal feasibility relations between simple primitives, e.g., \textit{hairy} is more feasible with \textit{cat} than with \textit{building} in the real world. Contextuality-dependence represents the contextual variance in images, e.g., \textit{cat} shows diverse appearances under the state of \textit{dry} and \textit{wet}. We design Semantic Attention (SA) and generative Knowledge Disentanglement (KD) to learn the dependence of feasibility and contextuality, respectively. SA captures semantics in compositions to alleviate impossible predictions, driven by the visual similarity between simple primitives. KD disentangles images into unbiased feature representations, easing contextual bias in predictions.
Moreover, we complement the current compositional probability model with feasibility and contextuality in a compatible format.
Finally, we conduct comprehensive experiments to analyze and validate the superior or competitive performance of our model, Semantic Attention and knowledge Disentanglement guided Simple Primitives (SAD-SP), on three widely-used benchmark OW-CZSL datasets.
\end{abstract}

}

\maketitle

\IEEEdisplaynontitleabstractindextext

%
\IEEEpeerreviewmaketitle

\IEEEraisesectionheading{\section{Introduction}\label{sec:introduction}}

%
%
%
%

\IEEEPARstart{M}{any} datasets exhibit long-tailed distribution, i.e., a large number of classes have few or even no prior instances~\cite{zhang2017range,zhao2021hierarchical,socher2013zero,zhang2021bag,samuel2021generalized,yan2022semantics,pourpanah2022review,romera2015embarrassingly}. Insufficient data become a bottleneck limiting the universality of deep learning~\cite{wang2019survey,wang2020generalizing,hospedales2021meta,ruis2021independent,xian2018zero,ramesh2021zero,changpinyo2016synthesized}. Comparatively, humans can intuitively identify non-existent concepts (e.g., \textit{canvas tree}), once humans understand the underlying primitives (e.g., \textit{canvas} and \textit{tree}). Inspired by this, recent works~\cite{nagarajan2018attributes,purushwalkam2019task,misra2017red,li2020symmetry,atzmon2020causal} propose a new learning paradigm named Compositional Zero-Shot Learning (CZSL). CZSL models images as compositions of primitive state and object concepts~\cite{Karthik_2022_CVPR,mancini2022learning,karthik2021revisiting,huynh2020compositional}. It aims to extract states and objects in seen images, transferring knowledge from seen to unseen, thereby recognizing unseen state-object compositions without training. For example, given images of \textit{canvas shoe} and \textit{brown tree}, machines can learn simple primitives of \textit{shoe} and \textit{brown}, thus directly recognizing the unseen composition of \textit{brown shoe} in images.

CZSL is challenging due to the context-dependent appearances~\cite{purushwalkam2019task,misra2017red}. For example, \textit{small} scales differently for elephants and cats; cats look different when they are \textit{young} or \textit{old}.
In other words, states and objects lead to visual changes to each other. Simple primitives in images are entangled visually~\cite{yang2022decomposable}. Most current works~\cite{nagarajan2018attributes,li2020symmetry,yang2022decomposable,li2022siamese,zhang2022learning,saini2022disentangling} view states or objects as the bases. They simulate the visual changes on bases caused by contextuality, inferring the possible entangled embedding. Thus, they can learn the visual embedding specific to each composition, helping distinguish unseen images. Despite the good performance of this strategy in Closed-World CZSL (CW-CZSL), it tends to largely degrade in Open-World CZSL (OW-CZSL)~\cite{Karthik_2022_CVPR,mancini2021open,mancini2022learning}. CW-CZSL provides the possible compositions as priors to simplify the inference, while OW-CZSL anonymizes these compositions and predicts compositions in the entire compositional space. For example, C-GQA~\cite{mancini2021open} (one of the benchmark datasets) contains 413 states and 674 objects. CW-CZSL performs inferences on a limited search space of 6,515 compositions (5,592 seen compositions for training and 923 unseen compositions for testing), approximately 2\% of the entire compositional space (278,362 possible compositions) in OW-CZSL. The large cardinality of possible compositions largely impairs the discrimination ability of compositional embedding~\cite{Karthik_2022_CVPR}.

Some OW-CZSL methods~\cite{li2020symmetry,mancini2021open,mancini2022learning,Karthik_2022_CVPR} exploit the unequal feasibility to ease impairment of the large learning cardinality.
These methods assign each composition a feasibility score to represent the probability of its existence, and they discard the impossible compositions. For example, \textit{canvas tree} is much less likely to exist in the world than \textit{canvas shoe}. Models can assign a large feasibility to \textit{canvas shoe} but a small feasibility to \textit{canvas tree}, eliminating \textit{canvas tree} by a feasibility threshold to make models focus on real-world images.
Current methods usually calculate the feasibility relying on external semantic knowledge~\cite{Karthik_2022_CVPR} or pairwise comparisons of label embedding~\cite{li2020symmetry,mancini2021open,mancini2022learning}. It may be infeasible when datasets lack relevant semantic information or have a large number of simple primitives.
Some other works~\cite{misra2017red,Karthik_2022_CVPR,karthik2021revisiting} decompose the composition into simple primitives. They assume that states/objects follow independent probability distributions. The compositional probability distribution can be the joint distribution of simple primitives, i.e., $p(\mathit{composition}|\mathit{image})=p(\mathit{state}|\mathit{image})p(\mathit{object}|\mathit{image})$. Since the cardinality of the simple primitives is much smaller than the cardinality of the compositions, the discrimination ability can be better preserved. However, the assumption of independent distributions ignores contextual and feasible relations between simple primitives~\cite{Karthik_2022_CVPR}, leading to biased predictions.

\begin{figure}[!t]
    \centering
    \includegraphics[width=\linewidth]{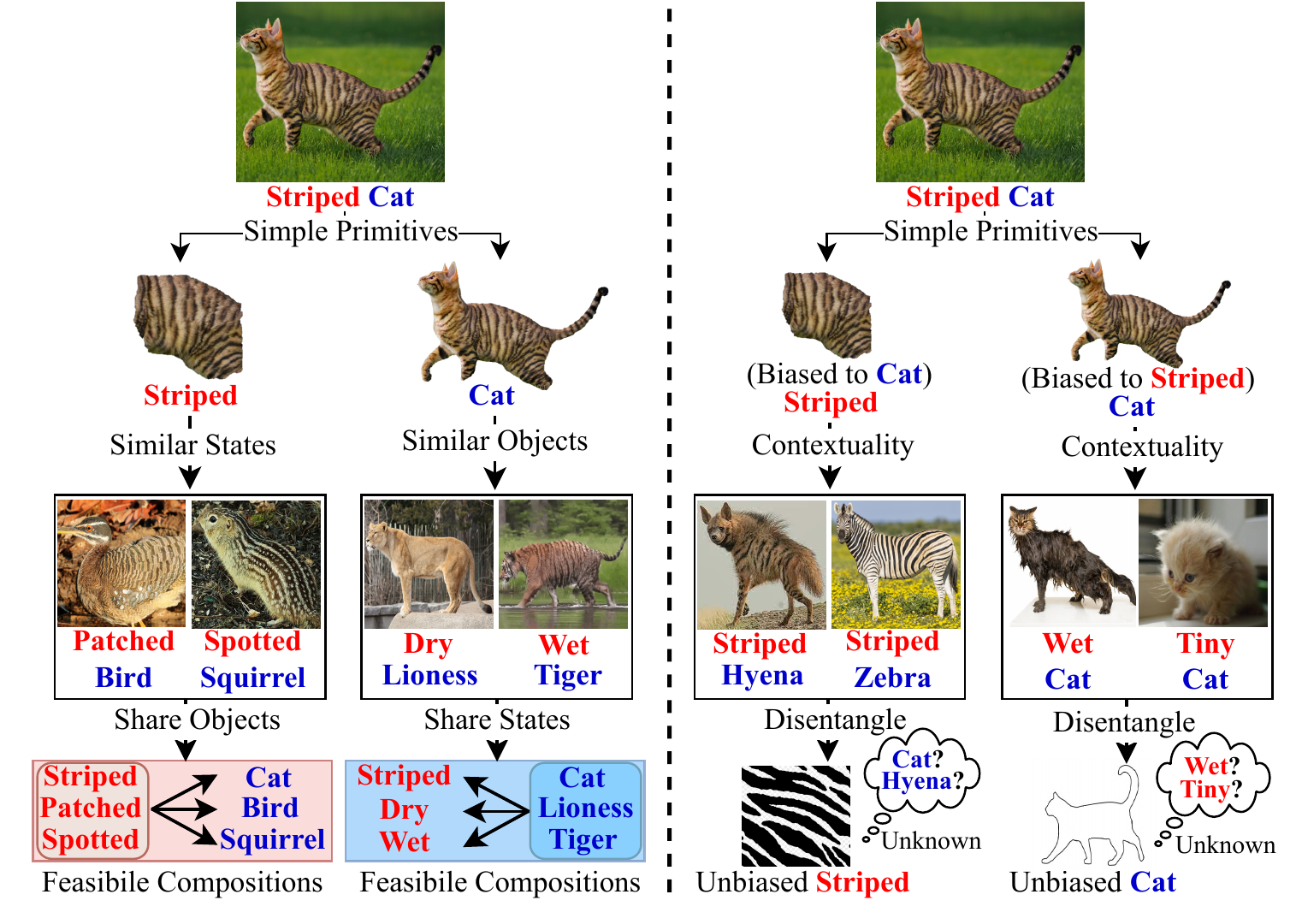}
    \caption{Illustration of our ideas on tackling feasibility (left) and contextuality (right) issues. 
    \textbf{Left}: We consider new compositions are highly feasible if they can be composed by similar states (or objects) sharing their feasible objects (or states). 
    For example, 
    visual similarity between \textit{Cat} and \textit{Tiger} can produce a highly feasible composition of \textit{Striped} \textit{Tiger}. 
    \textbf{Right}: We consider feature representations unbiased when they do not contain information of other simple primitives. For example, the disentangled \textit{Cat} is unbiased to contextuality because we cannot tell \textit{Tiny} or \textit{Wet} from it. 
    }
    \label{intro_motiv}
\end{figure}

Motivated by the above issues, we introduce two ideas to enable independent simple primitives to be compatible with learning contextuality and feasibility. First, inspired by the finding that the feasibility of similar states is shared among similar object~\cite{mancini2021open}, we assume that similar visual primitives share similar compositional feasibility. As shown in Figure~\ref{intro_motiv} (left), \textit{striped} and \textit{cat} are two simple primitives learned from the image independently. Similar visual patterns (e.g., striped and spotted) and species (e.g., cat and tiger) tend to be exchangeable in the compositions. Therefore, we may exchange simple primitives in seen compositions
to compose some new possible compositions, e.g., spotted cat and dry tiger, which tend to have higher feasibility than other unseen compositions. 
Taking advantage of the point that all objects and states are seen in OW-CZSL, we can utilize the semantics in existing compositions to infer the feasibility distribution of all possible compositions. 
Second, since it may be infeasible to simulate all visual changes of contextuality on states/objects, we tackle contextuality by disentangling the entangled visuals into disentangled feature representations. In Figure~\ref{intro_motiv} (right), \textit{striped} is not identical on different species, e.g., \textit{cat} and \textit{hyena}. Models might classify striped patterns into different classes due to the difference.
We propose to disentangle images into unbiased feature representations that do not contain visual changes of contextuality. For example, we learn disentangled representations of \textit{striped} not containing information of \textit{cat} or \textit{hyena}. Then, \textit{striped} patterns on different species can better be recognized as a unified class. Similarly, we can learn a unified \textit{cat} under different states.

In this paper, we implement our ideas in a unified model called Semantic Attention and knowledge Disentanglement guided Simple Primitives (SAD-SP). SAD-SP consists of three branches: Simple Primitive (SP), Semantic Attention (SA), and Knowledge Disentanglement (KD). SP follows the conventional practice of learning independent probability distributions for simple primitives. SA is a simple attention module driven by the inherent visual similarity between primitives to learn the shared compositional feasibility. KD is fulfilled by a generative adversarial network. We propose a distributional loss to supervise the generative network to exclude contextual information for unbiased representations. Both SA and KD have two parallel networks corresponding to the state and object, respectively. Unlike conventional methods taking SP as the final predictions, we use probabilistic revisions of the feasibility and the contextuality to complement the independent probability distributions.

Our contributions can be summarized as follows:
1) We propose two compatible ideas to complement conventional SP learning in a unified model. The unified model combines the learning of feasibility, contextuality, and independent simple primitives in a probabilistic format, which keeps a small learning cardinality. 
2) We design the parallel SA and KD to learn state/object-specific information. Our SA can infer compositional feasibility without external knowledge or pairwise similarity comparison. We propose a new distributional loss to enable KD to learn unbiased representation. 
3) We conduct extensive experiments to show the outperformance or the competitive performance of SAD-SP on benchmark datasets in the open-world setting. We show detailed quantitative and qualitative analyses to prove the model's effectiveness in improving compositional predictions.

\section{Related Work}
\subsection{Compositional Zero-shot Learning}
Compositional learning~\cite{naeem20223d,yang2022decomposable,hou2020visual,hou2021detecting,kato2018compositional,cho2017compositional,materzynska2020something} aims to enable models to learn simple primitives in images. Compositional zero-shot learning further recognizes unseen compositions in images composed of seen simple primitives. Simple primitives usually consist of two types of primitive concepts, e.g., states and objects~\cite{misra2017red,wei2019adversarial,li2022siamese,anwaar2021contrastive}. CZSL works~\cite{purushwalkam2019task,mancini2022learning,misra2017red,karthik2021revisiting,nagarajan2018attributes} attempt to learn discriminative visual representations of concepts to conduct classification. These methods can be divided into two main categories: compositional classification and simple primitive classification. Methods~\cite{misra2017red,naeem2021learning,nagarajan2018attributes,li2020symmetry} based on compositional classification project visual information into a common embedding space to recognize compositions directly. For example, Misra et al.~\cite{misra2017red} configure a specific compositional classifier for each composition to project inputs into the corresponding compositional space for classification. Nagarajan et al.~\cite{nagarajan2018attributes} view states as operators and perform all possible transformations on objects to build the embedding space for compositional classification.
Contrary to directly classifying compositions, simple primitive methods predict primitive concepts independently and then construct the compositional probability distribution for classification jointly~\cite{lu2016visual,misra2017red,karthik2021revisiting,Karthik_2022_CVPR}. These methods assume that states/objects follow independent probability distributions. They consider the compositional probability distribution as the joint probability distribution of objects and states, i.e., the product of predicted probabilities of objects and states. In other words, the problem of recognizing compositions is decomposed into recognizing two simple primitives independently. For example, Misra et al.~\cite{misra2017red} apply two independent classifiers to separately predict simple primitives and take the product of primitives as the probability of compositions. Karthik et al.~\cite{karthik2021revisiting} further propose a scalar bias to ease biased predictions towards seen compositions and enhance performance. Our method is similar to simple primitive methods but considers simple primitives' contextuality- and feasibility-dependence. We use contextuality and feasibility as revisions to complement state-object dependence in probabilities of simple primitives. The attention is threshold-free, which uses feasibility to calibrate probabilities directly. Thus, we do not need post-processing to select a threshold by hand engineering.

\subsection{Open-world Compositional Zero-shot Learning}
Open-world compositional zero-shot learning is more realistic and challenging than conventional closed-world setting. In the open-world setting, no priors about possible unseen compositions are given. CW-CZSL works tend to be less effective due to the large cardinality of possible compositions~\cite{nagarajan2018attributes,purushwalkam2019task,atzmon2020causal,misra2017red,Karthik_2022_CVPR}. However, some works in CW-CZSL can still be enlighting to tackle the large cardinality. For example, Li et al.~\cite{li2020symmetry} mask impossible compositions by computing pair probability based on the distance between state and object categories. Xu et al.~\cite{xu2021relation} adopt a key-query-based attention mechanism to capture the correlation between primitive concepts in a graph to pass messages selectively. In OW-CZSL, similar to \cite{li2020symmetry}, Mancini et al.~\cite{mancini2021open,mancini2022learning} propose to utilize the graph structure to model the dependence between state, object, and compositions. In this way, some infeasible compositions can be eliminated during inference. However, these feasibility computation is based on pairwise similarity comparison. Pairwise comparison tends to be less effective on datasets with large numbers of states and objects~\cite{Karthik_2022_CVPR}. Another way to reduce composition cardinality is simple primitives~\cite{misra2017red,karthik2021revisiting,Karthik_2022_CVPR}. The paradigm of simple primitives independently predicting states and objects can naturally reduce cardinality. Karthik et al.~\cite{Karthik_2022_CVPR} further improve simple primitives by estimating the feasibility of each composition through external knowledge to eliminate impossible state-object pairs. Different from \cite{Karthik_2022_CVPR,xu2021relation}, our work estimates feasibility with attention mechanism and implies feasibility in the form of probability to enhance simple primitive learning.


\subsection{Knowledge Disentanglement in Zero-shot Learning}
Visual variance in images caused by contextuality is a common issue in zero-shot learning~\cite{lian2022robust,jhoo2021collaborative,gabbay2021image,li2022zero,tong2019hierarchical,fan2022contrastive,chen2021semantics,li2021generalized}. Plenty of previous works have been done to utilize knowledge disentanglement to learn invariant or unbiased visual representations. For example, Chen et al.~\cite{chen2021semantics} and Li et al.~\cite{li2021generalized} use a conditional VAE to disentangle images into semantic-consistent and semantic-unrelated latent vectors. Li et al.~\cite{li2021disentangled} generatively disentangle inputs into sub-features to obtain simple independent hierarchical features. Knowledge disentanglement can also be applied in disentangling graphs. Geng et al.~\cite{geng2022disentangled} disentangle knowledge graphs into ontology embeddings to capture fine-grained semantic information. In compositional zero-shot learning, Yang et al.~\cite{yang2022decomposable} disentangle images into states and objects based on the causal effects in compositions. Saini et al.~\cite{saini2022disentangling} apply a state/object affinity network to disentangle the same states or objects from image pairs contrastively. Zhang et al.~\cite{zhang2022learning} reconsider CZSL as
an out-of-distribution generalization problem and use domain alignment in the gradient level to disentangle images into object-invariant and attribute-invariant features. In this paper, we propose a novel generative network to disentangle images based on the probability distributions of classes, which can be effective in the OW-CZSL setting.

\section{Method}
\subsection{Problem Formulation}
Given an image dataset $X=\{x_{i}:i\in[1,N]\}$ with compositional labels $Y=\{y_{i}:i\in[1,N]\}$, $N$ denotes the dataset size, and $(x_{i},y_{i})$ is an image $x_{i}$ with the corresponding compositional label $y_{i}$. CZSL~\cite{mancini2021open} aims to recognize state-object compositions $y_{i}=(s,o)$ based on the given sets of states $s\in S$ and objects $o\in O$. In this paper, we focus on the challenging OW-CZSL. OW-CZSL recognizes both seen and unseen object-state compositions from the whole compositional space, i.e., $Y_{\mathit{space}}=S\times O$. OW-CZSL divides $X$ into a training set $(X_{tr},Y_{tr})$ and a testing set $(X_{ts},Y_{ts})$. The training set is used to train the model, where $Y_{tr}$ contains all possible states and objects but not all possible compositions. The testing set consists of both seen compositions and unseen compositions during the training stage. Note that seen compositions have no intersection with unseen compositions.

\subsection{Simple Primitives in CZSL}
Following the SP baselines in CZSL~\cite{Karthik_2022_CVPR,misra2017red}, we predict the probabilities of states/objects independently and take the product of their probabilities as the final results to recognize compositions. Our SP consists of two parts: extractor and classifier. The extractor learns the feature representations of the input. The classifier predicts the state/object probabilities based on the learned feature representations.

Formally, given an image $x$, we have an extractor $f_{e}: X\rightarrow Z$ embedding the image $x$ to a feature representation $z$. Then, a state classifier $f_{s}: Z\rightarrow \Delta_{S}$ and an object classifier $f_{o}: Z\rightarrow \Delta_{O}$ predicts the state probability $p_{s}\in [0,1]^{1\times |S|}$ and object probability $p_{o}\in[0,1]^{1\times |O|}$, respectively, where $|S|$ and $|O|$ denote the number of states and objects in the dataset. $p_{s}$ and $p_{o}$ are vectors that $\sum_{k=1}^{|S|} p_{s_{k}}=1,\sum_{j=1}^{|O|} p_{o_{j}}$. $p_{s_{k}}$ and $p_{o_{j}}$ represent the prediction probability of the $k^{th}$ state and the $j^{th}$ object, respectively. Then, the composition probability for each state-object pair can be calculated by the probabilities of independent predictions $p_{(s_{k},o_{j})}=p_{s_{k}}\times p_{o_{j}}$.

We use the cross-entropy loss to train the independent predictions of states and objects as follows:

\begin{equation}
    \begin{split}
        \min_{<f_{e},f_{s},f_{o}>}\mathcal{L}_{sp}&=\sum_{i=1}^{N}\mathcal{L}_{ce}(f_{s}(f_{e}(x_{i})),s_{i})+\mathcal{L}_{ce}(f_{o}(f_{e}(x_{i})),o_{i})\\
        &=-\sum_{i=1}^{N}\log f_{s}(z^{s}_{i},s_{i})+\log f_{o}(z^{o}_{i},o_{i})
    \end{split}
\end{equation}
where $\mathcal{L}_{ce}$ denotes the cross-entropy loss; $z_{i}= f_{e}(x_{i})$; $z^{s}_{i}$ and $z^{}_{i}$ denote the corresponding branch for predictions in $z_{i}$; $s_{i}$ and $o_{i}$ are the corresponding ground-truth state and object labels for $x_{i}$. $f_{s}(z^{s}_{i},s_{i})=p_{s_{i}}$ is the probability assigned to state $s_{i}$ of $x_{i}$; $f_{o}(z^{o}_{i},o_{i})=p_{o_{i}}$ is the probability assigned to object $o_{i}$ of $x_{i}$.

\begin{figure*}[h]
    \centering
    \includegraphics[width=\textwidth]{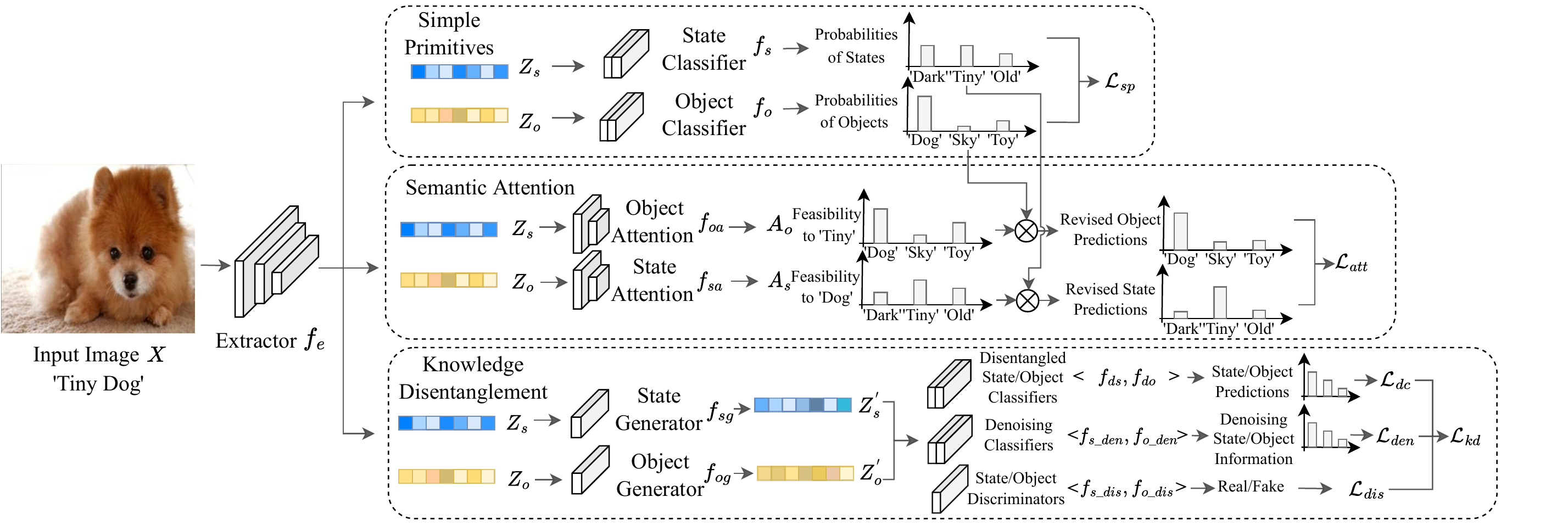}
    \caption{Model overview. SAD-SP consists of three components: Simple Primitives (SP), Semantic Attention (SA), and Knowledge Disentanglement (KD). The three components use the shared feature representations $Z=\{Z_{s},Z_{o}\}$ extracted by an extractor $f_{e}$. (1) We adopt two classifiers ($f_{s}$ and $f_{o}$) to predict states and objects in the SP stage independently, which is supervised by loss $\mathcal{L}_{sp}$. (2) Semantic attention uses object attention $f_{oa}$ and state attention $f_{sa}$ to learn the probabilities (i.e., $A_{o}$ and $A_{s}$) of compositions to represent semantic relations of states and objects. For example, `Tiny' has a high probability of being a composition with `Dog' but a low probability with `Sky'. Then, we use loss $\mathcal{L}_{att}$ to supervise SA to revise and improve the SP predictions. (3) We use state generator $f_{sg}$ and object generator $f_{og}$ to obtain the disentangled state $Z_{s}^{'}$ and object $Z_{o}^{'}$, respectively. The generators are trained in an adversarial way with a pair of parallel classifiers and discriminators.
    Specifically, $<f_{ds},f_{do}>$ is for state/object predictions $\mathcal{L}_{dc}$; $<f_{s\_den},f_{o\_den}>$ is for denoising state/object information $\mathcal{L}_{den}$; $<f_{s\_dis},f_{o\_dis}>$ is for distinguishing real/fake features $\mathcal{L}_{dis}$; $\mathcal{L}_{kd}$ is the overall loss for KD.}
    \label{model_verview}
\end{figure*}

\subsection{Simple Primitives with Feasibility and Contextuality}
In this section, we introduce the details of our proposed semantic attention and knowledge disentanglement to tackle feasible and contextual dependence between simple primitives. The model overview of SAD-SP is shown in Figure~\ref{model_verview}. 

\subsubsection{Simple Primitives with Dependence}
The conventional SP ignores the dependence between states, objects, and compositions, while the dependence can be effective in composition recognition~\cite{mancini2022learning,naeem2021learning}. Thus, we assume conventional SP as a basis for representing the predictions based on the superficial semantics in images. We let the dependence between simple primitives be the high-level semantic information influencing the basic probability. Then, for an arbitrary composition $(s_{k},o_{j})$, our new compositional probability can be modeled as follows:
\begin{equation}\label{new_model_of_p_s_o}
\resizebox{0.9\hsize}{!}{$\begin{aligned}
        p_{(s_{k},o_{j})}&=p_{s_{k}}^{'}\times p_{o_{j}}^{'}\\
&=(p_{s_{k}}+p_{f}(s_{k}|o_{j})+p_{c}(s_{k}))\times(p_{o_{j}}+p_{f}(o_{j}|s_{k})+p_{c}(o_{j}))\\
\end{aligned}$}
\end{equation}
where $p_{s_{k}}^{'}$ and $p_{o_{j}}^{'}$ denote the probability of simple primitive with dependence; $p_{s_{k}}$ and $p_{o_{j}}$ denote the independent state and object probability; $p_{f}$ and $p_{c}$ represent the probabilistic revision provided by feasibility- and contextuality-dependence, respectively. 

We consider each word has its own semantic space. The feasibility of compositions may differ for the state and the object. For example, the feasibility of \textit{white polar bear} differs for \textit{white} and \textit{polar bear}: when we describe polar bears, we may use the state of \textit{white} frequently; however, when we use \textit{white}, we may frequently associate it with common white things in daily life, e.g., snow, but not the rare polar bears. Therefore, we assume that the predictions of states and objects are correct. Then, we can use $p_{f}(s_{k}|o_{j})$ and $p_{f}(o_{j}|s_{k})$ to represent the conditioned feasibility of $(s_{k},o_{j})$ for the state $s_{k}$ and the object $o_{j}$, respectively.

\subsubsection{Semantic Attention for Feasibility}
We learn state/object-conditioned compositional feasibility via semantic attention following the idea: unseen compositions composed by exchanging similar simple primitives in seen compositions are more feasible than other unseen compositions. 
We design an attention module to implement our idea, driven by the inherent similarity of feature representation during training: similar visual patterns tend to be embedded as similar feature representations~\cite{purushwalkam2019task}. In other words, similar simple primitives tend to have similar feature representations. Thus, we can use parallel attention to learn conditioned semantic relations for states and objects, respectively. For example, given compositions \textit{wet tiger} and \textit{small cat}, if we use attention to learn from objects to help predict states, the attention will build strong feasibility from \textit{tiger} to \textit{wet} and from \textit{cat} to \textit{small}. Due to the similarity between \textit{tiger} and \textit{cat}, when the object is \textit{tiger} or \textit{cat}, the attention tends to assign high feasibility to state \textit{wet} and state \textit{small}, which is consistent with our idea.

Formally, given a feature representation $Z=\{Z_{s},Z_{o}\}$, we let $Z_{s}$ and $Z_{o}$ be the features from the corresponding branches for states and objects, respectively. We have an object-conditioned attention for state $f_{sa}: Z_{o}\rightarrow A_{s}$ and a state-conditioned attention for object $f_{oa}: Z_{s}\rightarrow A_{o}$ to learn the feasibility between states and objects. We use the Sigmoid function to let $a^{s}_{i}\in (0,1)^{1\times |S|}$ and $a^{o}_{i}\in (0,1)^{1\times |O|}$ be vectors that have the same size as the state/object set. Each element in the vector represents the strength of feasibility between the corresponding state and object. The larger element means the larger probability of the composition of the state and object existing in the dataset. We use the learned semantics to improve the search space of SP by fusing the attention map with the predictions via the element-wise product, i.e., $p_{f}(s_{k}|o_{j})=a^{s}_{k}\times p_{s_{k}}$, $p_{f}(o_{j}|s_{k})=a^{o}_{j}\times p_{o_{j}}$. Then, we minimize the following loss for training:

\begin{equation}
\resizebox{0.9\hsize}{!}{$\begin{split}
    \mathcal{L}_{att}&=\sum_{i=1}^{N}\mathcal{L}_{ce}(f_{sa}(z^{o}_{i})\otimes f_{s}(z^{s}_{i},s_{i}))+\mathcal{L}_{ce}(f_{oa}(z^{s}_{i})\otimes f_{o}(z^{o}_{i},o_{i}))\\
    &=-\sum_{i=1}^{N}\log a^{s}_{\sigma(s_{i})}\otimes  f_{s}(z^{s}_{i},s_{i})+\log a^{o}_{\sigma(o)}\otimes  f_{o}(z^{o}_{i},o_{i})
\end{split}$}
\end{equation}
where $\sigma(s_{i})$ and $\sigma(o_{i})$ return the location of the ground-truth labels in attention, respectively; $a^{s}_{\sigma(s_{i})}\otimes  f_{s}(z^{o}_{i},s_{i})=a^{s}_{\sigma(s_{i})} f_{s}(z^{s}_{i},s_{i})+ f_{s}(z^{s}_{i},s_{i})$; $z^{s}_{i}$ and $z^{o}_{i}$ denote the state and object branch of $z_{i}$; $a^{s}$ and $a^{o}$ denote the state and object attention maps.

$\mathcal{L}_{att}$ trains attention maps $a^{s}$ and $a^{o}$ to find the suitable probabilities to improve the final predictions. The attention maps learn semantic relations from existing images to measure the composition probabilities dependent on states and objects. 
This shifts the model focus to possible compositions to refine the search space. Then, the probabilistic revision of objects/states can be viewed as the auxiliary classification information learned from the states/objects to help enhance the original predictions, which are guided by the semantic relations between simple primitives. Since we view the feasibility as auxiliary information to make predictions focused on possible compositions, we do not need a post-processing threshold value to help eliminate the impossible compositions~\cite{Karthik_2022_CVPR}.

\subsubsection{Knowledge Disentanglement for Contextuality} 
Given the entangled feature representations $Z=\{Z_{s}, Z_{o}\}$, we disentangle feature representations to obtain unbiased visual information in an adversarial way. 
We apply two parallel generators, discriminators, and classifiers to disentangle the state/object. We use a state generator $f_{sg}: Z_{s}\rightarrow Z_{s}^{'}$ and an object generator $f_{og}: Z_{o}\rightarrow Z_{o}^{'}$ to generate the disentangled state representation and object representation, respectively. We propose two principles to ensure representations of states and objects are disentangled: (\rom{1}) The disentangled feature representations should be able to predict the target classes accurately but not the non-target classes. For example, the disentangled state representations should only retain the state information but not the object information. (\rom{2}) The disentangled feature representations should be `real' as the originally learned feature representations. This means that the discriminators cannot distinguish between the disentangled and the original feature representations. The principle (\rom{1}) ensures that the model can learn unbiased feature representations. The principle (\rom{2}) is a regularization requirement that regularizes the feature representation to preserve visual information rather than generating noise not matching the original images.

To fulfill the principle (\rom{1}), we design disentangled classifiers and denoising classifiers to supervise the model learning required information. The disentangled state classifier $f_{ds}: \{Z_{s},Z_{s}^{'}\}\rightarrow \Delta_{S}$ and the disentangled object classifier $f_{do}: \{Z_{o},Z_{o}^{'}\}\rightarrow \Delta_{O}$ supervise the disentangled feature representations (i.e., $z^{s'}_{i}$ and $z^{o'}_{i}$) carrying the same classification information as the original feature representations (i.e., $z^{s}_{i}$ and $z^{o}_{i}$). We use denoising classifiers $f_{s\_den}: \{Z_{s},Z_{s}^{'}\}\rightarrow \{\textbf{1}_{O},U_{O}\}$ and $f_{o\_den}: \{Z_{o},Z_{o}^{'}\}\rightarrow \{\textbf{1}_{S},U_{S}\}$ to disentangle and denoise the non-target information in the disentangled feature representations, where $U$ denotes the uniform distribution and $\textbf{1}$ denotes the one-hot distribution. We use the uniform distribution, i.e., $U_{S}$ and $U_{O}$, to represent `disentangled', which means that the feature representations cannot distinguish classes. All the classes have the same probability of $\frac{1}{|S|}$ or $\frac{1}{|U|}$ in predictions. We let the one-hot distributions, i.e., $\textbf{1}_{S}$ and $\textbf{1}_{O}$, represent 'non-disentangled', which means that the feature representations still can accurately assign a probability of 1 to the ground-truth class of state or object. 
Then, we can use cross-entropy loss and mean square error loss to learn disentangled features as follows:
\begin{equation}
    \resizebox{0.9\hsize}{!}{$\begin{gathered}
        \min_{<f_{d},f_{g}>} \mathcal{L}_{dc}=-\mathbb{E}_{z_{i}\sim Z\cup Z^{'}}\sum_{i}^{N}\log f_{ds}(z^{s}_{i},s_{i})+\log f_{do}(z^{o}_{i},o_{i})\\
    \end{gathered}$}
\end{equation}

\begin{equation}
\resizebox{0.9\hsize}{!}{$\begin{gathered}
    \max_{f_{den}}\mathcal{L}_{den}=-\mathbb{E}_{z_{i}\sim Z\cup Z^{'}}\frac{1}{|S|}\sum ||f_{o\_den}(z^{o}_{i},s_{i})-\textbf{1}_{S}(z_{i})||^{2}\\
    -\mathbb{E}_{z_{i}\sim Z\cup Z^{'}}\frac{1}{|O|}\sum ||f_{s\_den}(z^{s}_{i},o_{i})-\textbf{1}_{O}(z_{i})||^{2}
\end{gathered}$}
\end{equation}

\begin{equation}
\begin{gathered}
    \min_{f_{g}}\mathcal{L}_{den}=\mathbb{E}_{z_{i}\sim Z^{'}}\frac{1}{|S|}\sum ||f_{o\_den}(z^{o}_{i},s_{i})-U_{S}(z_{i})||^{2}\\
    +\mathbb{E}_{z_{i}\sim Z^{'}}\frac{1}{|O|}\sum ||f_{s\_den}(z^{s}_{i},o_{i})-U_{O}(z_{i})||^{2}
\end{gathered}
\end{equation}
where $Z^{'}=\{Z_{s}^{'},Z_{o}^{'}\}$; $f_{d}=\{f_{ds},f_{do}\}$ denotes the set of disentangled classifiers; $f_{g}=\{f_{sg},f_{og}\}$ denotes the generators to disentangle knowledge; $f_{den}=\{f_{s\_den},f_{o\_den}\}$ denotes the classifiers to denoise non-target knowledge.
$f_{ds}(z^{s}_{i},s_{i})$ and $f_{do}(z^{o}_{i},o_{i})$ denote class probability of $f_{ds}$/$f_{do}$ assigned to the ground-truth label $s_{i}$/$o_{i}$ of the input. $f_{o\_den}(z^{o}_{i},s_{i})$ and $f_{s\_den}(z^{s}_{i},o_{i})$ denote the predicted state/object probability distribution from feature representations of object/state; $\textbf{1}_{S/O}(z_{i})$ denotes a one-hot distribution that only the ground-truth state/object label of $z_{i}$ is one; $U_{S/O}(z_{i})$ denotes a uniform distribution that each class has an equal probability in the predictions of states/objects for $z_{i}$.

$\mathcal{L}_{dc}$ optimizes the disentangled classifier to learn the target classification information from both generated and original feature representation. Then, $\mathcal{L}_{dc}$ supervises the generator to generate features with the classification information of the target state/object classes.

$\mathcal{L}_{den}$ is a 2-stage min-max loss function. $\mathcal{L}_{den}$ first optimizes $\{f_{s\_den},f_{o\_den}\}$ to be capable of precisely predicting non-target classes. Then, $\mathcal{L}_{den}$ supervises generators to disentangle feature representations, updating generators to denoise the non-target information in feature representations, i.e., denoising object/state information in $z^{s}_{i}$/$z^{o}_{i}$.

To fulfill the principle (\rom{2}), we use discriminators $f_{s\_dis}: \{Z_{s},Z_{s}^{'}\}\rightarrow \{0,1\}$ and $f_{o\_dis}: \{Z_{o},Z_{o}^{'}\}\rightarrow \{0,1\}$ to distinguish real and fake feature representations of states and objects, where 0 denotes `fake' and 1 denotes `real'. We consider the generated feature representations as `fake' and the original feature representations as `real'. The goal is to optimize the generator to fool the discriminators. We optimize the generator and discriminator with cross-entropy loss in a min-max format as follows:

\begin{equation}
\resizebox{0.9\hsize}{!}{$\begin{gathered}
    \max_{f_{dis}}\mathcal{L}_{dis}=\mathbb{E}_{z_{i}\sim Z}[\log f_{s\_dis}(z^{s}_{i},s_{i})+\log f_{o\_dis}(z^{o}_{i},o_{i})]\\
    +\mathbb{E}_{z_{i}\sim Z^{'}}[\log (1-f_{s\_dis}(z^{s}_{i},s_{i}))+\log (1-f_{o\_dis}(z^{o}_{i},o_{i}))]\\
    \min_{f_{g}}\mathcal{L}_{dis}=-\mathbb{E}_{z_{i}\sim Z^{'}}[\log f_{s\_dis}(z^{s}_{i},s_{i})+\log f_{o\_dis}(z^{o}_{i},o_{i})]
\end{gathered}$}
\end{equation}
where $f_{dis}=\{f_{s\_dis},f_{o\_dis}\}$ is the discriminators; $Z^{'}$ the generated features by $f_{g}$.

$\mathcal{L}_{dis}$ first optimizes the discriminators to accurately distinguish the generated and original feature representations. It then supervises generators to fool the discriminators, which can regularize the disentangled feature representations to mimic the original representations to preserve important visual information in the original images.

Finally, we can summarize the distributional loss $\mathcal{L}_{kd}$ of knowledge disentanglement in a min-max format as follows:
\begin{equation}
    \min_{<f_{d},f_{g}>}\max_{<f_{dis},f_{den}>}\mathcal{L}_{kd}=\mathcal{L}_{dc}+\mathcal{L}_{den}+\mathcal{L}_{dis}
\end{equation}

We can obtain disentangled state and object representations generated by $f_{sg}$ and $f_{og}$ for unbiased predictions. We view the unbiased predictions as a probabilistic revision to shift model focus to the predictions without contextual variance, i.e., given the feature representation $\{z^{s}_{i},z^{o}_{i}\}$ of an image, for an arbitrary composition $(s_{k},o_{j})$, $p_{c}(s_{k})=f_{ds}(f_{sg}(z^{s}_{i}),s_{k})$ and $p_{c}(o_{j})=f_{do}(f_{og}(z^{o}_{i}),o_{j})$.

\subsubsection{Summary}

The overall loss function of SAD-SP is as follows:
\begin{equation}
    \mathcal{L}_{SAD\textit{-}SP}=\mathcal{L}_{sp}+\mathcal{L}_{att}+\mathcal{L}_{kd}
\end{equation}

For an arbitrary image $x_{i}$, the predictions in the inference stage can be summarized:

\begin{equation}
\argmax_{(s_{k},o_{j})}p_{(s_{k},o_{j})}=p_{s_{k}}^{'}\times p_{o_{j}}^{'}
\end{equation}
\begin{equation}
p_{s_{k}}^{'}= \gamma_{1}f_{s}(z^{s}_{i},s_{k})+\gamma_{2}a^{s}_{k} f_{s}(z^{s}_{i},s_{k})+ \gamma_{3}f_{ds}(f_{sg}(z^{s}_{i}),s_{k})
\end{equation}
\begin{equation}
p_{o_{j}}^{'}= \gamma_{1}f_{o}(z^{o}_{i},o_{j})+\gamma_{2}a^{o}_{j} f_{o}(z^{o}_{i},o_{j})+ \gamma_{3}f_{do}(f_{og}(z^{o}_{i}),o_{j})
\end{equation}
where $\gamma_{1}$, $\gamma_{2}$, $\gamma_{3}$ are hyper-parameters for tuning SP predictions to different datasets by grid search; $\gamma_{1}+\gamma_{2}+\gamma_{3}=1$.

\section{Experiment}
\subsection{Experiment Settings and Implementation Details}
\textbf{Datasets.} We evaluate our proposed SAD-SP on three commonly used OW-CZSL benchmark datasets, i.e., UT-Zappos dataset~\cite{yu2014fine,yu2017semantic}, MIT-States dataset~\cite{isola2015discovering} and C-GQA dataset~\cite{naeem2021learning}. The UT-Zappos dataset is a small dataset only containing shoes. It consists of 12 different shoe types and 16 footwear materials. We view each shoe type as an object class and each footwear material as a state class. Different from the dataset made of a single object, MIT-States and C-GQA datasets are two large datasets containing diverse objects (e.g., buildings and animals) and states (e.g., shapes and colors). MIT-States dataset has 245 different object classes and 115 state types. C-GQA is a compositional version of Stanford GQA dataset~\cite{hudson2019gqa} with 674 object categories and 413 state categories. In the open-world scenario, we follow the standard split in previous works~\cite{Karthik_2022_CVPR,purushwalkam2019task}: images ($\sim$23k images) from 83 compositions out of 192 possible compositions ($\sim$43\%) are seen for training on the UT-Zappos dataset; images ($\sim$30k images) from 1,262 compositions out of 28,175 possible compositions ($\sim$4\%) are used for training on the MIT-States dataset; images ($\sim$27k images) from 5,592 compositions out of 27,8362 possible compositions ($\sim$2\%) are presented for training on the C-GQA dataset. About 3k images from 36 compositions (18 seen and 18 unseen compositions), 13k images from 800 compositions (400 seen and 400 unseen compositions), and 5k images from 1,811 compositions (888 seen and 923 unseen compositions) are selected for testing on UT-Zappos, MIT-States, and C-GQA, respectively.

\begin{table*}[h]
\centering
\caption{Results on three OW-CZSL benchmark datasets. We evaluate models by the best accuracy of seen classes (S), unseen classes (U), harmonic mean (HM), and the best score of area under the curve (AUC). ${\text{ff}}$ represents the fixed backbone. \textbf{*} denotes using external knowledge besides ImageNet. The best results are in \textbf{bold}. The second-best results are \underline{underlined}. }
\begin{tabular}{|c|cccc|cccc|cccc|}
\toprule
          & \multicolumn{4}{c}{MIT-States}                                                              & \multicolumn{4}{|c|}{UT-Zappos}                                                                                       & \multicolumn{4}{c|}{C-GQA}                                                                                               \\\cmidrule{2-13}
 \multirow{-2}{*}{Model}   & S                           & U                          & HM                         & AUC                        & S                           & U                           & HM                          & AUC                         & S                            & U                           & HM                          & AUC                          \\\midrule
 Compositional Embedding&&&&&&&&&&&&\\
LE+~\cite{misra2017red}       & 14.2                        & 2.5                        & 2.7                        & 0.3                        & 60.4                        & 36.5                        & 30.5                        & 16.3                        & 19.2                         & 0.7                         & 1.0                         & 0.08                         \\
AoP~\cite{nagarajan2018attributes}       & 16.6                        & 5.7                        & 4.7                        & 0.7                        & 50.9                        & 34.2                        & 29.4                        & 13.7                        & -                            & -                           & -                           & -                            \\
TMN~\cite{purushwalkam2019task}       & 12.6                        & 0.9                        & 1.2                        & 0.1                        & 55.9                        & 18.1                        & 21.7                        & 8.4                         & -                            & -                           & -                           & -                            \\
SymNet~\cite{li2020symmetry}    & 21.4                        & 7.0                        & 5.8                        & 0.8                        & 53.3                        & 44.6                        & 34.5                        & 18.5                        & 26.7                         & 2.2                         & 3.3                         & 0.43                         \\

CompCos$^{\mathit{CW}}$\textbf{*}~\cite{mancini2021open} & 25.3                        & 5.5                        & 5.9                        & 0.9                        & 59.8                        & 45.6                        & 36.3                        & 20.8                        & 28.0                         & 1.0                         & 1.6                         & 0.20                         \\
CompCos\textbf{*}~\cite{mancini2021open}   & 25.4                        & \textbf{10.0}                       & \textbf{8.9}                        & \textbf{1.6}                        & 59.3                        & 46.8                        & 36.9                        & 21.3                        & 28.4                         & 1.8                         & 2.8                         & 0.39                         \\

CGE$_{\text{ff}}$\textbf{*}~\cite{naeem2021learning}     & \underline{29.6}                        & 4.0                        & 4.9                        & 0.7                        & 58.8                        & 46.5                        & 38.0                        & 21.5                        & 28.3                         & 1.3                         & 2.2                         & 0.30                         \\
CGE\textbf{*}~\cite{naeem2021learning}       & \textbf{32.4}                        & 5.1                        & 6.0                        & 1.0                        & 61.7                        & 47.7                        & 39.0                        & 23.1                        & \textbf{32.7}                         & 1.8                         & 2.9                         & 0.47                         \\

\midrule
Simple Primitives&&&&&&&&&&&&\\
VisProd~\cite{misra2017red}   & 20.9                        & 5.8                        & 5.6                        & 0.7                        & 54.6                        & 42.8                        & 36.9                        & 19.7                        & 24.8                         & 1.7                         & 2.8                         & 0.33                         \\
VisProd$_{\text{ff}}$++~\cite{karthik2021revisiting}&24.6&6.7&6.6&1.0&58.3&47.1&39.3&22.8&27.2&2.1&3.3&0.46\\
VisProd++~\cite{karthik2021revisiting}&28.1&7.5&7.3&1.2&\underline{62.5}&51.5&41.8&26.5&28.0&2.8&4.5&0.75\\
KG-SP$_{\text{ff}}$\textbf{*}~\cite{Karthik_2022_CVPR}   & 23.4                        & 7.0                        & 6.7                        & 1.0                        & 58.0                        & 47.2                        & 39.1                        & 22.9                        & 26.6                         & 2.1                         & 3.4                         & 0.44                         \\
KG-SP\textbf{*}~\cite{Karthik_2022_CVPR}     & 28.4                        & 7.5                        & 7.4                        & 1.3                        & 61.8                        & \underline{52.1}                        & \underline{42.3}                        & \underline{26.5}                        & 31.5                       & 2.9                         & 4.7                         & 0.78                         \\\midrule
SAD-SP$_{\text{ff}}$ & 27.4                        & 7.0                        & 7.1                        & 1.2                        & 60.6                        & 46.0                        & 40.0                        & 23.1                        & \underline{32.5}                        & \underline{3.2}                        & \underline{5.1}                        & \underline{0.86}                         \\
SAD-SP    & 29.1 &  \underline{7.6} & \underline{7.8} &  \underline{1.4} &  \textbf{63.1} &  \textbf{54.7} &  \textbf{44.0} &  \textbf{28.4} &  31.0 &  \textbf{3.9} &  \textbf{5.9} &  \textbf{1.002}\\
\bottomrule
\end{tabular}
\label{exp_overall_model}
\end{table*}

\textbf{Metrics and Benchmarks.} For a fair comparison, we follow the standard evaluation protocol in previous works ~\cite{Karthik_2022_CVPR,mancini2021open}. We evaluate the seen accuracy (S), the unseen accuracy (U), the harmonic mean (HM) of seen and unseen accuracy, and the area under the curve (AUC) under the generalized compositional zero-shot setting. In generalized zero-shot learning, models are required to conduct inferences on both seen and unseen classes to show their generalization ability. However, the predictions are often biased to seen classes significantly. We thus use a bias calibration that is commonly used in zero-shot learning~\cite{xu2020attribute,xie2019attentive,chao2016empirical} and compositional zero-shot learning~\cite{Karthik_2022_CVPR,purushwalkam2019task} to ease the biased predictions. Same as the common operation of bias calibration in CZSL~\cite{Karthik_2022_CVPR,purushwalkam2019task}, we use varying constants as bias terms to measure the best S, U, HM, and AUC scores (in \%) as our final results.

We compare our methods with 9 representative methods: LE+~\cite{misra2017red}, AoP~\cite{nagarajan2018attributes}, TMN~\cite{purushwalkam2019task}, SymNet~\cite{li2020symmetry}, CGE~\cite{naeem2021learning}, CompCos~\cite{mancini2021open}, VisProd~\cite{misra2017red}, VisProd++~\cite{karthik2021revisiting}, and KG-SP~\cite{Karthik_2022_CVPR}. LE+, AoP, TMN, SymNet, CGE, and CompCos are methods based on the compositional classification that predict compositional labels in a shared embedding space. VisProd, VisProd++, and KG-SP are similar to our SAD-SP, which takes the probabilistic product of simple primitives to infer the compositions. More specifically, LE+ and VisProd are two baseline models; CGE and CompCos${\mathit{CW}}$ enhance LE+ by building labels in graph embedding based on word2vec+fastext~\cite{mikolov2013distributed,bojanowski2017enriching}; VisProd++ and KG-SP improve VisProd by advanced network architecture and bias calibration; AoP, SymNet, and TMN use generative network or modulator to simulate the contextual variance in images; CompCos and KG-SP use pairwise similarity in graph structure or external semantic knowledge to inject feasibility into compositional predictions. Specially, we take the fixed backbone version of compared methods (e.g., CGE$_{\text{ff}}$) to evaluate the improvement derived from the end-to-end training, in which case the backbone will not be fine-tuned and the input can be viewed as the extracted representations of the backbone. We omit an unfair State-Of-The-Art (SOTA) comparison with Co-CGE~\cite{mancini2022learning} that is a combination of CGE and CompCos.

\textbf{Implement details.} We use the standard network architecture, i.e., ResNet18~\cite{he2016deep}, as the backbone. We adopt a three-layer Fully Connected Network (FCN) structure for each module in SAD-SP, namely $f_{s}$, $f_{o}$, $f_{oa}$, $f_{sa}$, $f_{den}$, $f_{dis}$, and $f_{d}$. The three-layer structure consists of a two-layer FCN for extracting feature representations and a one-layer FCN for making predictions. In particular, we consider $f_{sg}$ and $f_{og}$ as the shared two-layer extractor of $f_{den}$, $f_{dis}$ and $f_{d}$, which extracts the disentangled feature representations from the input. Then, we use the Adam optimizer~\cite{kingma2014adam} with the same parameters for the three datasets. We use the default parameters of Adam optimizer to train different modules in SAD-SP with the same weight decay of 5.0e-5~\cite{krogh1991simple} but different learning rates. We train the backbone with a learning rate of 5.0e-6, $f_{den}$ $f_{dis}$ with a learning rate of 1.0e-2, and other modules in SAD-SP with a learning rate of 5.0e-5. We use a set of balanced $\gamma$ parameters $\gamma_{1}=0.7$, $\gamma_{2}=0.25$, and $\gamma_{3}=0.05$ by searching $\gamma_{2}$, $\gamma_{3}$ in the range of 0.05 to 0.5 with a step of 0.05 on three datasets instead of searching delicate hyper-parameters for each dataset. The experiments are implemented in PyTorch~\cite{paszke2019pytorch} and NVIDIA TITAN X with CUDA 11.0~\cite{cuda}. 

\subsection{Open-World Compositional Zero-Shot Learning}

Among methods predicting simple primitives in Table~\ref{exp_overall_model}, our proposed SAD-SP achieves the best performance on all criteria for OW-CZSL. On MIT-States, UT-Zappos, and C-GQA, SAD-SP consistently outperforms SOTA methods in HM and AUC. Compared with the best SOTA method (i.e., KG-SP), SAD-SP relatively improves HM by 5.4\% (7.4 vs 7.8), 4.0\% (42.3 vs 44.0), 25.5\% (4.7 vs 5.9) and increases AUC by 7.7\% (1.3 vs 1.4), 7.2\% (26.5 vs 28.4), 28.5\% (0.78 vs 1.002), respectively. Note that KG-SP applies external knowledge to train a concept network to eliminate some impossible compositions. The consistent improvement of HM and AUC indicates that knowledge disentanglement and semantic attention in SAD-SP can effectively provide meaningful semantic information related to feasibility and contextuality in compositional predictions, which can be as effective as external semantic knowledge. SAD-SP achieves the best S and U scores except SAD-SP$_{\text{ff}}$ obtaining the highest S on C-GQA. In other words, when applying extremely large or small bias terms, constraining the output range into seen or unseen compositions, our method can have the best ability to fit seen compositions and to be generalized to unseen compositions.

SAD-SP still achieves the highest scores on UT-Zappos and C-GQA compared to methods that project the input into a shared embedding space. Compared with the best SOTA methods (i.e. CGE and SymNet), SAD-SP relatively improves AUC by 12.8\% (39.0 vs 44.0), 78.8\% (3.3 vs 5.9) and 22.9\% (23.1 vs 28.4), 111.3\% (0.47 vs 1.002). On both datasets, models based on combined embeddings tend to perform worse than models that predicted simple primitives. On MIT-States, SAD-SP shows the second-best performance, and CompCos exhibits the best performance. However, improvements of CompCos are dataset-specific. CompCos and CGE are two graph embedding methods. CompCos outperforms CGE on MIT-States but underperforms CGE on other datasets. This suggests that CompCos injecting feasibility computed by pairwise cosine similarity of label embeddings can reduce the inherent label noise on MIT-States~\cite{atzmon2020causal}, but this cosine feasibility may harm or not affect model performance on other datasets. On the contrary, our model has robustness on different datasets.

Compared with baselines (i.e., LE+ and VisProd), balancing seen and unseen compositions can boost the performance, e.g., VisProd++. Learning contextuality or feasibility can also enhance model performance, e.g., AoP and KG-SP. Moreover, the models using external semantic knowledge (i.e., CompCos, CGE, and KG-SP) tend to have better performance than other methods following the same manner of recognizing compositions. The external knowledge is usually used as semantic experts who project words of labels into graph embedding or help eliminate infeasible compositions. The effectiveness of external knowledge suggests that a significant factor in recognizing compositions precisely is to capture the semantic relations between primitive concepts and their combinations. Different from these models, SAD-SP does not rely on external knowledge except the commonly used backbone trained on ImageNet, but SA and KD can still improve the model ability as effectively as the external knowledge. In other words, SA and KD can learn more semantic information that remained unused yet informative than conventional methods. We also exhibit the performance without end-to-end training, i.e., CGE$_{\text{ff}}$, VisProd$_{\text{ff}}$++, KG-SP$_{\text{ff}}$, and SAD-SP$_{\text{ff}}$. We can observe that SAD-SP$_{\text{ff}}$ defeats other methods in HM and AUC on three datasets. It indicates that SA and KD have the strongest discriminative ability when they are applied at the top of the same backbones.


\subsection{Feasibility and Contextuality Analysis}
\subsubsection{Ablation Study}

\begin{table}[t!]
\centering
\caption{AUC and HM scores of module-level variants. ${\text{ff}}$ denotes the fixed backbone. The best results in this table for each dataset (row) are in \textbf{bold}. The best results among Table~\ref{exp_each_module}-\ref{exp_each_module_ablation_hm} for each dataset (row) are further colored in \textcolor{red}{red}.
}
\begin{tabular}{|cl|cccc|}
\hline
\multicolumn{2}{|c|}{}                           & \multicolumn{4}{c|}{AUC}                                                                                                                                                 \\ \cline{3-6} 
\multicolumn{2}{|c|}{\multirow{-2}{*}{Datasets}} & \multicolumn{1}{c|}{SP}     & \multicolumn{1}{c|}{SA-SP}          & \multicolumn{1}{c|}{KD-SP}                                  & SAD-SP                                 \\ \hline
\multicolumn{2}{|c|}{MIT-States}                 & \multicolumn{1}{c|}{1.270}  & \multicolumn{1}{c|}{1.274}          & \multicolumn{1}{c|}{1.275}                                  & {\color[HTML]{FF0000} \textbf{1.362}}  \\ \hline
\multicolumn{2}{|c|}{MIT-States$_{\text{ff}}$}   & \multicolumn{1}{c|}{1.141}  & \multicolumn{1}{c|}{1.141}          & \multicolumn{1}{c|}{1.159}                                  & \color{red}{\textbf{1.190}}                         \\ \hline
\multicolumn{2}{|c|}{UT-Zappos}                  & \multicolumn{1}{c|}{28.425} & \multicolumn{1}{c|}{28.166}        & \multicolumn{1}{c|}{{\color[HTML]{FF0000} \textbf{28.453}}} & 28.406                                 \\ \hline
\multicolumn{2}{|c|}{UT-Zappos$_{\text{ff}}$}  & \multicolumn{1}{c|}{22.875} & \multicolumn{1}{c|}{22.768}         & \multicolumn{1}{c|}{22.985}                                 & {\color[HTML]{FF0000} \textbf{23.079}} \\ \hline
\multicolumn{2}{|c|}{C-GQA}                      & \multicolumn{1}{c|}{0.964}  & \multicolumn{1}{c|}{0.975}          & \multicolumn{1}{c|}{\textbf{1.006}}                         & 1.002                                  \\ \hline
\multicolumn{2}{|c|}{C-GQA$_{\text{ff}}$}        & \multicolumn{1}{c|}{0.851}  & \multicolumn{1}{c|}{\textbf{0.876}} & \multicolumn{1}{c|}{0.862}                                  & 0.856                                  \\ \hline
\multicolumn{2}{|c|}{}                           & \multicolumn{4}{c|}{HM}                                                                                                                                                  \\ \cline{3-6} 
\multicolumn{2}{|c|}{\multirow{-2}{*}{Datasets}} & \multicolumn{1}{c|}{SP}     & \multicolumn{1}{c|}{SA-SP}          & \multicolumn{1}{c|}{KD-SP}                                  & SAD-SP                                 \\ \hline
\multicolumn{2}{|c|}{MIT-States}                 & \multicolumn{1}{c|}{7.714}  & \multicolumn{1}{c|}{7.737}          & \multicolumn{1}{c|}{7.818}  & \textcolor{red}{\textbf{7.844}}                                  \\ \hline
\multicolumn{2}{|c|}{MIT-States$_{\text{ff}}$}   & \multicolumn{1}{c|}{7.003}  & \multicolumn{1}{c|}{6.949}          & \multicolumn{1}{c|}{7.115}                                  & {\color[HTML]{FF0000} \textbf{7.146}}  \\ \hline
\multicolumn{2}{|c|}{UT-Zappos}                  & \multicolumn{1}{c|}{43.441} & \multicolumn{1}{c|}{43.950}         & \multicolumn{1}{c|}{{\color[HTML]{FF0000} \textbf{44.068}}} & 43.990                                 \\ \hline
\multicolumn{2}{|c|}{UT-Zappos$_{\text{ff}}$}  & \multicolumn{1}{c|}{39.746} & \multicolumn{1}{c|}{39.886}         & \multicolumn{1}{c|}{39.887}                                 & {\color[HTML]{FF0000} \textbf{40.030}} \\ \hline
\multicolumn{2}{|c|}{C-GQA}                      & \multicolumn{1}{c|}{5.795}  & \multicolumn{1}{c|}{5.803}          & \multicolumn{1}{c|}{\textbf{6.086}}                         & 5.877                                  \\ \hline
\multicolumn{2}{|c|}{C-GQA$_{\text{ff}}$}        & \multicolumn{1}{c|}{5.108}  & \multicolumn{1}{c|}{\textbf{5.232}} & \multicolumn{1}{c|}{5.136}                                  & 5.140                                  \\ \hline
\end{tabular}
\label{exp_each_module}
\end{table}

\begin{table}[h]
\centering
\caption{AUC scores of branch-level variants. ${\text{ff}}$ denotes the fixed backbone. The best results in this table for each dataset (column) are in \textbf{bold}. The best results among Table~\ref{exp_each_module}-\ref{exp_each_module_ablation_hm} for each dataset (column) are further colored in \textcolor{red}{red}.
}
\begin{tabular}{|cl|c|c|c|}
\hline
\multicolumn{2}{|c|}{}                          &                                            &                                           &                                       \\
\multicolumn{2}{|c|}{\multirow{-2}{*}{Disable}} & \multirow{-2}{*}{MIT-States}               & \multirow{-2}{*}{UT-Zappos}               & \multirow{-2}{*}{C-GQA}               \\ \hline
\multicolumn{2}{|c|}{$p_{f}(s|o)$}              & 1.281                                      & 28.117                          & 0.992                                 \\ \hline
\multicolumn{2}{|c|}{$p_{f}(o|s)$}              & 1.272                                      & \textbf{28.265}                           & 0.994                                 \\ \hline
\multicolumn{2}{|c|}{$p_{c}(s)$}                & 1.279                                      & 26.287                                    & {\color[HTML]{FF0000} \textbf{1.045}} \\ \hline
\multicolumn{2}{|c|}{$p_{c}(o)$}                & 1.283                            & 27.851                                    & 0.994                                 \\ \hline
\multicolumn{2}{|c|}{$p_{f}(s|o)$\&$p_{c}(s)$}  & 1.279                                      & 25.819                                    & 1.008                                 \\ \hline
\multicolumn{2}{|c|}{$p_{f}(s|o)$\&$p_{c}(o)$}  & \textbf{1.285}                             & 27.737                                    & 0.967                                 \\ \hline
\multicolumn{2}{|c|}{$p_{f}(o|s)$\&$p_{c}(s)$}  & 1.270                                      & 26.070                                    & 1.013                                 \\ \hline
\multicolumn{2}{|c|}{$p_{f}(o|s)$\&$p_{c}(o)$}  & 1.268                                      & 27.799                                    & 1.000                                 \\ \hline
\multicolumn{2}{|c|}{}                          &                                            &                                           &                                       \\
\multicolumn{2}{|c|}{\multirow{-2}{*}{Disable}} & \multirow{-2}{*}{MIT-States$_{\text{ff}}$} & \multirow{-2}{*}{UT-Zappos$_{\text{ff}}$} & \multirow{-2}{*}{C-GQA$_{\text{ff}}$} \\ \hline
\multicolumn{2}{|c|}{$p_{f}(s|o)$}              & 1.153                                      & 22.857                                    & 0.837                                 \\ \hline
\multicolumn{2}{|c|}{$p_{f}(o|s)$}              & 1.164                                      & \textbf{23.000}                           & {\color[HTML]{FF0000} \textbf{0.877}} \\ \hline
\multicolumn{2}{|c|}{$p_{c}(s)$}                & 1.162                             & 22.299                                    & 0.848                                 \\ \hline
\multicolumn{2}{|c|}{$p_{c}(o)$}                & \textbf{1.166}                             & 22.774                           & 0.869                                 \\ \hline
\multicolumn{2}{|c|}{$p_{f}(s|o)$\&$p_{c}(s)$}  & 1.152                                      & 22.051                                    & 0.814                                 \\ \hline
\multicolumn{2}{|c|}{$p_{f}(s|o)$\&$p_{c}(o)$}  & 1.144                                      & 22.465                                    & 0.874                                 \\ \hline
\multicolumn{2}{|c|}{$p_{f}(o|s)$\&$p_{c}(s)$}  & 1.153                                      & 22.181                                    & 0.854                                 \\ \hline
\multicolumn{2}{|c|}{$p_{f}(o|s)$\&$p_{c}(o)$}  & 1.165                                      & 22.678                                    & 0.867                                 \\ \hline
\end{tabular}
\label{exp_each_module_ablation_auc}
\end{table}

\begin{table}[h]
\centering
\caption{HM scores of branch-level variants. ${\text{ff}}$ denotes the fixed backbone. The best results in this table for each dataset (column) are in \textbf{bold}. The best results among Table~\ref{exp_each_module}-\ref{exp_each_module_ablation_hm} for each dataset (column) are further colored in \textcolor{red}{red}.
} 
\begin{tabular}{|cl|c|c|c|}
\hline
\multicolumn{2}{|c|}{}                          &                                            &                                           &                                       \\
\multicolumn{2}{|c|}{\multirow{-2}{*}{Disable}} & \multirow{-2}{*}{MIT-States}               & \multirow{-2}{*}{UT-Zappos}               & \multirow{-2}{*}{C-GQA}               \\ \hline
\multicolumn{2}{|c|}{$p_{f}(s|o)$}              & 7.766                                      & \textbf{43.874}                           & 5.837                                 \\ \hline
\multicolumn{2}{|c|}{$p_{f}(o|s)$}              & 7.746                                      & 43.823                                    & 5.921                                 \\ \hline
\multicolumn{2}{|c|}{$p_{c}(s)$}                & 7.793                                      & 42.767                                    & {\color[HTML]{FF0000} \textbf{6.094}} \\ \hline
\multicolumn{2}{|c|}{$p_{c}(o)$}                & \textbf{7.836}                             & 43.483                                    & 5.982                                 \\ \hline
\multicolumn{2}{|c|}{$p_{f}(s|o)$\&$p_{c}(s)$}  & 7.784                                      & 42.545                                    & 5.863                                 \\ \hline
\multicolumn{2}{|c|}{$p_{f}(s|o)$\&$p_{c}(o)$}  & 7.827                                      & 43.266                                    & 5.833                                 \\ \hline
\multicolumn{2}{|c|}{$p_{f}(o|s)$\&$p_{c}(s)$}  & 7.719                                      & 42.583                                    & 5.915                                 \\ \hline
\multicolumn{2}{|c|}{$p_{f}(o|s)$\&$p_{c}(o)$}  & 7.709                                      & 43.647                                    & 5.912                                 \\ \hline
\multicolumn{2}{|c|}{}                          &                                            &                                           &                                       \\
\multicolumn{2}{|c|}{\multirow{-2}{*}{Disable}} & \multirow{-2}{*}{MIT-States$_{\text{ff}}$} & \multirow{-2}{*}{UT-Zappos$_{\text{ff}}$} & \multirow{-2}{*}{C-GQA$_{\text{ff}}$} \\ \hline
\multicolumn{2}{|c|}{$p_{f}(s|o)$}              & 7.005                                      & 39.452                                    & 5.163                                 \\ \hline
\multicolumn{2}{|c|}{$p_{f}(o|s)$}              & 7.033                                      & 39.661                                    & {\color[HTML]{FF0000} \textbf{5.237}} \\ \hline
\multicolumn{2}{|c|}{$p_{c}(s)$}                & \textbf{7.111}                             & 39.418                                    & 5.138                                 \\ \hline
\multicolumn{2}{|c|}{$p_{c}(o)$}                & 7.059                                      & \textbf{39.819}                           & 5.136                                 \\ \hline
\multicolumn{2}{|c|}{$p_{f}(s|o)$\&$p_{c}(s)$}  & 7.044                                      & 38.876                                    & 5.046                                 \\ \hline
\multicolumn{2}{|c|}{$p_{f}(s|o)$\&$p_{c}(o)$}  & 6.998                                      & 39.429                                    & 5.129                                 \\ \hline
\multicolumn{2}{|c|}{$p_{f}(o|s)$\&$p_{c}(s)$}  & 7.081                                      & 39.173                                    & 5.119                                 \\ \hline
\multicolumn{2}{|c|}{$p_{f}(o|s)$\&$p_{c}(o)$}  & 7.072                                      & 39.424                                    & 5.117                                 \\ \hline
\end{tabular}
\label{exp_each_module_ablation_hm}
\end{table}

In this section, we conduct ablation studies from a module-level and branch-level perspective, analyzing the effects of feasibility and contextuality on performance. Module-level studies reveal the effects of modules, i.e. SA and KD, on SP predictions; branch-level studies further analyze the detailed effects of SA and KD on model performance within branches (i.e., states and objects). Since VisProd++~\cite{karthik2021revisiting} in Table~\ref{exp_overall_model} can be viewed as the baseline of our model without any proposed modules or losses, we only disable modules or branches during the inference to demonstrate their effects. Then, SP in Table~\ref{exp_each_module} can represent a baseline with the enhanced network parameters enhanced by our losses. For a fair comparison, we use the same $\gamma$ parameters as Table \ref{exp_overall_model}, i.e., $\gamma_{1}=0.7$, $\gamma_{2}=0.25$ and $\gamma_{3}=0.05$.

\textbf{Module-level Analysis}.
~In Table \ref{exp_each_module}, we disable SA and KD to exhibit the AUC and HM results of module-level variants, namely SP, SA-SP, and KD-SP. SP represents the results of the enhanced SP branch by $\mathcal{L}_{att}$; SA-SP and KD-SP are variants disabling the modules of KD and SA, respectively. Comparing SP with VisProd++, when not using end-to-end training, SP$_{\text{ff}}$ achieves similar performance on UT-Zappos, slightly increases the performance on MIT-States, but largely improves the performance on C-GQA. The improvements of SP$_{\text{ff}}$ are consistent with the improvements of SA-SP, suggesting that $\mathcal{L}_{att}$ significantly boosts our model on C-GQA. When using end-to-end training, SP with fine-tuned backbone can increase performance by 0.07/0.414, 1.925/1.641, and 0.214/1.295 in AUC/HM on MIT-States, UT-Zappos, and C-GQA, respectively. The improvements validate that $\mathcal{L}_{att}$ can help improve the learning ability of the backbone and the prediction layers.

Moreover, we can observe that SAD-SP achieves the best AUC scores across all variants on MIT-States, MIT-States$_{\text{ff}}$ and UT-Zappos$_{\text{ff}}$. It also achieves the highest HM scores in the corresponding datasets, indicating that both SA and KD can benefit the compositional recognition in a complementary way on MIT-States and UT-Zappos with or without end-to-end training. Compared with SP, KD-SP consistently improves model performance on different datasets, especially by a large margin on UT-Zappos and C-GQA (e.g., HM score: 43.441 vs 44.068 and 5.795 vs 6.086). The consistent improvement of KD indicates the effectiveness and universality of knowledge disentanglement on learning contextuality in OW-CZSL. SA-SP can increase the AUC and HM scores on C-GQA but may impair model performance on MIT-States and UT-Zappos. The demerit may be caused by the label noise on MIT-States and weak semantic feasibility relations between shoes and materials on UT-Zappos. Nonetheless, the feasibility information learned by SA can boost KD under most conditions.

\begin{figure*}[ht]
    \centering 
\hspace{-1mm}
\begin{subfigure}{0.44\textwidth}
  \includegraphics[width=\textwidth]{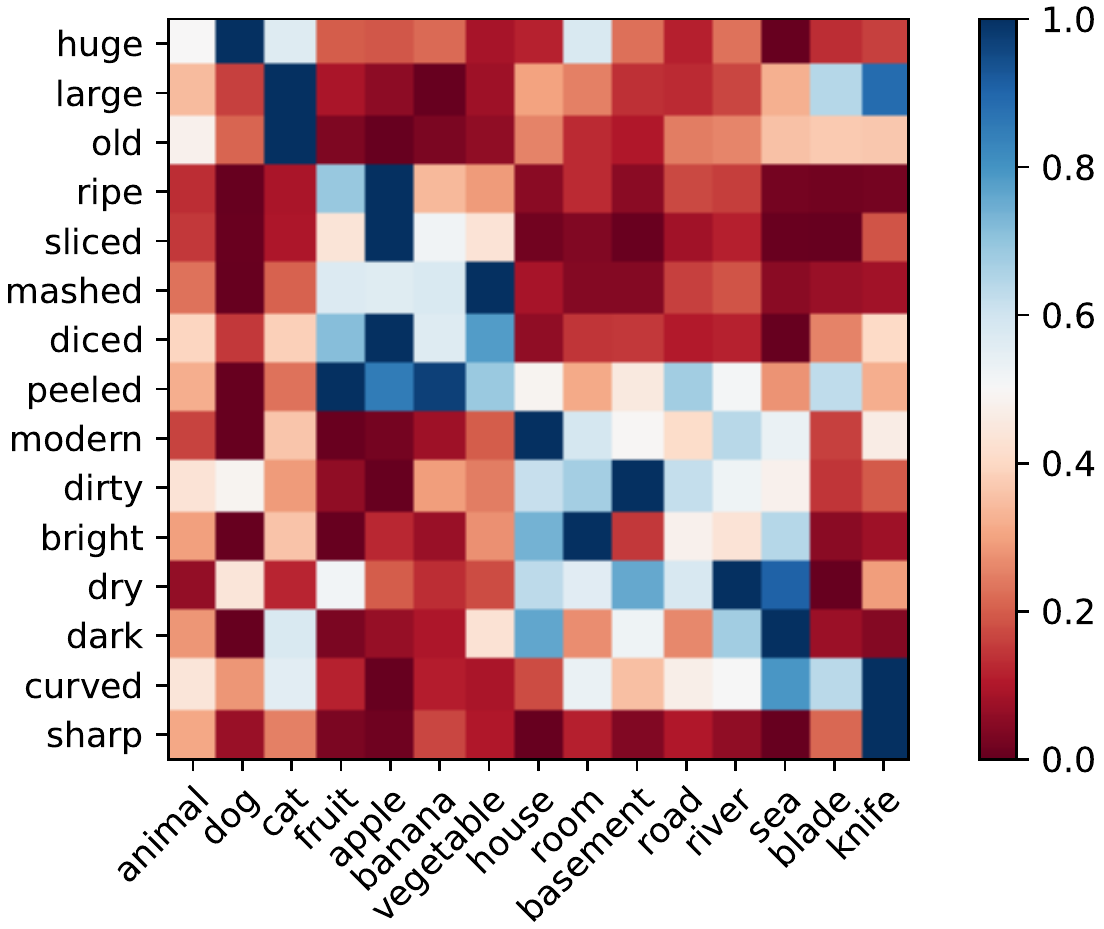}
    \centering
    \caption{Object attention (row) conditioned on the state.}
\end{subfigure}
\begin{subfigure}{0.44\textwidth}
  \includegraphics[width=\textwidth]{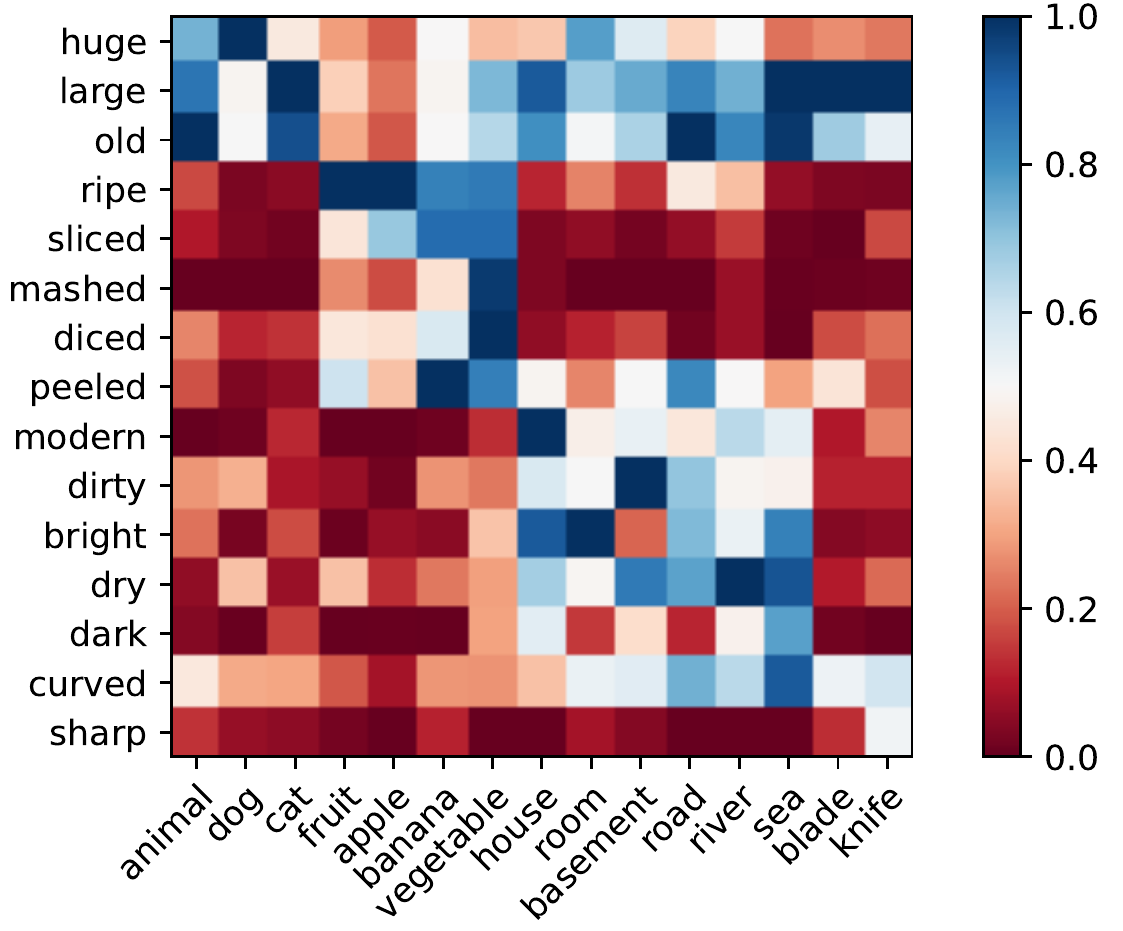}
    \centering
    \caption{State attention (column) conditioned on the object.}
\end{subfigure}
\caption{Visualization of weight distributions for (a) object attention and (b) state attention on MIT-States.}
\label{feature_embedding}
\end{figure*}
\textbf{Branch-level Analysis}. We exhibit AUC and HM scores by disabling branches within SAD-SP in Table \ref{exp_each_module_ablation_auc}-\ref{exp_each_module_ablation_hm}, respectively. $p_{f}(s|o)$ and $p_{c}(s)$ denote probabilistic revisions of feasibility and contextuality provided by semantic attention and knowledge disentanglement for the state branch. Similarly, $p_{f}(o|s)$ and $p_{c}(o)$ are the corresponding revisions for the object branch. On MIT-States and MIT-States$_{\text{ff}}$, we can observe that disabling any branch in SAD-SP will lead to a lower HM or AUC score. This indicates that both semantic attention and knowledge disentanglement in the state or object branch can provide complementary information for compositional recognition. Comparing feasibility and contextuality branches, we can observe that the learned feasibility (i.e., $p_{f}(o|s)$ and $p_{f}(s|o)$) are more informative than the learned contextuality (i.e., $p_{c}(o)$ and $p_{c}(s)$). Disabling $p_{f}(o|s)$ or $p_{f}(s|o)$ may cause severe information loss while removing contextuality causes the least information loss. For example, on MIT-States, disabling $p_{f}(o|s)$ results in decreasing AUC up to 0.090 while disabling $p_{f}(s|o)\&p_{c}(o)$ achieves the highest AUC increase, i.e., 0.004, compared with disabling $p_{f}(s|o)$. On MIT-States$_{\text{ff}}$, disabling $p_{f}(s|o)$ results in decreasing HM up to 0.141 while disabling $p_{f}(o|s)\&p_{c}(s)$ achieves the highest HM increase, i.e., 0.048, compared with disabling $p_{f}(o|s)$.

On the opposite, knowledge disentanglement plays a more important role than semantic attention on UT-Zappos. Eliminating contextuality will cause a significant performance decrease, especially state contextuality. AUC drops by up to 2.298/1.329 and HM declines by up to 0.819/0.612 with/without end-to-end training due to disabling $p_{c}(s)$, indicating that learning unbiased material representations is a key factor in recognizing shoe compositions precisely. Feasibility seems to be less effective on UT-Zappos because disabling state or object attention has a limited effect on the model performance. However, SA can still provide some complementary information to boost KD performance when not using end-to-end training.

Though SAD-SP outperforms SOTA methods on C-GQA, disabling $p_{c}(s)$ and $p_{f}(o|s)$ can further enhance SAD-SP to achieve better performance. For example, disabling knowledge disentanglement of state in the end-to-end training can relatively increase AUC by 4.3\% (1.002 vs 1.045) and HM by 3.7\% (5.877 vs 6.094); removing semantic attention of object in the non-end-to-end training relatively raises AUC by 2.5\% (0.856 vs 0.877) and HM by 1.9\% (5.140 vs 5.237). The improvements indicate that SAD-SP sometimes fails to learn the most balanced feasibility- and contextuality-dependence on C-GQA; however, SAD-SP still has a strong ability to capture the semantic information of dependence between simple primitives.


\begin{table*}[t]
\centering
\caption{Examples of Top-3 feasible compositions in SA on MIT-States and C-GQA datasets. \textbf{OW} and \textbf{UC} represent the learned feasible compositions existing in the Open-World (OW) compositional space or Unseen Compositional (UC) space. \textbf{GT} denotes the Ground-Truth (GT) unseen compositions in the testing set. Compositions existing in the training set are in \textit{italics}. The overlapped simple primitives are in \textbf{bold}.}

\begin{tabular}{|ccclcccl|}
\hline
\multicolumn{1}{|c|}{State}                     & \multicolumn{3}{c|}{Top-3 Feasible Objects}                                                    & \multicolumn{1}{c|}{Object}                    & \multicolumn{3}{c|}{Top-3 Feasible States}                                                 \\ \hline
\multicolumn{8}{|c|}{MIT-States}                                                                                                                                                                                                                                                             \\ \hline
\multicolumn{1}{|c|}{\multirow{3}{*}{straight}} & \multicolumn{1}{c|}{OW}     & \multicolumn{2}{c|}{\textit{road}, \textit{sword}, \textbf{blade}}                       & \multicolumn{1}{l|}{\multirow{3}{*}{~velvet}}   & \multicolumn{1}{c|}{OW}     & \multicolumn{2}{c|}{\textit{brushed}, \textit{crushed}, \textbf{wrinkled}}           \\ \cline{2-4} \cline{6-8} 
\multicolumn{1}{|c|}{}                          & \multicolumn{1}{c|}{UC}        & \multicolumn{2}{c|}{\textbf{blade}, bronze, \textbf{highway}}          & \multicolumn{1}{l|}{}                          & \multicolumn{1}{c|}{UC}        & \multicolumn{2}{c|}{\textbf{wrinkled}, creased, crumpled} \\ \cline{2-4} \cline{6-8} 
\multicolumn{1}{|c|}{}                          & \multicolumn{1}{c|}{GT}     & \multicolumn{2}{c|}{\textbf{blade}, \textbf{highway}, pool}             & \multicolumn{1}{l|}{}                          & \multicolumn{1}{c|}{GT}     & \multicolumn{2}{c|}{crumpled, \textbf{wrinkled}}          \\ \hline
\multicolumn{1}{|c|}{\multirow{3}{*}{squished}} & \multicolumn{1}{c|}{OW}     & \multicolumn{2}{c|}{\textit{sandwich}, \textit{tomato}, \textbf{bread}}                  & \multicolumn{1}{c|}{\multirow{3}{*}{blade}}    & \multicolumn{1}{c|}{OW}     & \multicolumn{2}{c|}{\textit{blunt}, large, \textbf{straight}}               \\ \cline{2-4} \cline{6-8} 
\multicolumn{1}{|c|}{}                          & \multicolumn{1}{c|}{UC}        & \multicolumn{2}{c|}{\textbf{bread},   \textbf{fish}, plate}            & \multicolumn{1}{c|}{}                          & \multicolumn{1}{c|}{UC}        & \multicolumn{2}{c|}{large, \textbf{straight}, \textbf{bent}}                \\ \cline{2-4} \cline{6-8} 
\multicolumn{1}{|c|}{}                          & \multicolumn{1}{c|}{GT}     & \multicolumn{2}{c|}{\textbf{bread},   bus, coin, \textbf{fish}, penny} & \multicolumn{1}{c|}{}                          & \multicolumn{1}{c|}{GT}     & \multicolumn{2}{c|}{\textbf{bent},   narrow, \textbf{straight}}    \\ \hline
\multicolumn{8}{|c|}{C-GQA}                                                                                                                                                                                                                                                                  \\ \hline
\multicolumn{1}{|c|}{\multirow{3}{*}{forested}} & \multicolumn{1}{c|}{OW}     & \multicolumn{2}{c|}{\textbf{hill},   \textit{mountain}, \textit{tree}}          & \multicolumn{1}{c|}{\multirow{3}{*}{mattress}} & \multicolumn{1}{c|}{OW}     & \multicolumn{2}{c|}{crumpled, \textbf{folded}, \textit{blue}}               \\ \cline{2-4} \cline{6-8} 
\multicolumn{1}{|c|}{}                          & \multicolumn{1}{c|}{UC}        & \multicolumn{2}{c|}{\textbf{hill},   cliff, forest}           & \multicolumn{1}{c|}{}                          & \multicolumn{1}{c|}{UC}        & \multicolumn{2}{c|}{crumpled, \textbf{folded}, carpeted}           \\ \cline{2-4} \cline{6-8} 
\multicolumn{1}{|c|}{}                          & \multicolumn{1}{c|}{GT}     & \multicolumn{2}{c|}{hillside, \textbf{hill}}                           & \multicolumn{1}{c|}{}                          & \multicolumn{1}{c|}{GT}     & \multicolumn{2}{c|}{red, soft, \textbf{folded}}                    \\ \hline
\multicolumn{1}{|c|}{\multirow{3}{*}{asian}}    & \multicolumn{1}{c|}{OW}     & \multicolumn{2}{c|}{\textit{boy}, building, \textbf{person}}                    & \multicolumn{1}{c|}{\multirow{3}{*}{tail}}     & \multicolumn{1}{c|}{OW}     & \multicolumn{2}{c|}{\textbf{blue},   hairy, \textit{long}}         \\ \cline{2-4} \cline{6-8} 
\multicolumn{1}{|c|}{}                          & \multicolumn{1}{c|}{UC}        & \multicolumn{2}{c|}{building, \textbf{person}, bleachers}              & \multicolumn{1}{c|}{}                          & \multicolumn{1}{c|}{UC}        & \multicolumn{2}{c|}{\textbf{blue},   hairy, worn}         \\ \cline{2-4} \cline{6-8} 
\multicolumn{1}{|c|}{}                          & \multicolumn{1}{c|}{GT}     & \multicolumn{2}{c|}{\textbf{person}}                          & \multicolumn{1}{c|}{}                          & \multicolumn{1}{c|}{GT}     & \multicolumn{2}{c|}{\textbf{blue},   silver, orange}      \\ \hline
\end{tabular}

\label{semantics_attention_example}
\end{table*}


\subsubsection{Feasibility Distribution of Semantic Attention}
In this section, we analyze the feasibility (i.e., weight) distributions of SA, showing the learned semantic relations in compositional feasibility from a comparative perspective. We first accumulate the instance-specific object and state attention according to their ground-truth labels, obtaining the dataset-level attention map. Then, we use min-max normalization to normalize the attention map conditioned on the state or object to learn the comparative relations.

Formally, given an arbitrary input, let the $i^{\mathit{th}}$ object $o_{i}$ and the $j^{\mathit{th}}$ state $s_{j}$ be the ground-truth labels of the input, we accumulate attention to learn a matrix $\mathcal{M}$ representing the weight distribution as follows:
\begin{equation}
\begin{gathered}
      \mathcal{M}_{i}=Softmax(\mathcal{M}_{i}+a^{o})\\
    \mathcal{M}^{T}_{j}=Softmax(\mathcal{M}^{T}_{j}+a^{s})
\end{gathered}
\end{equation}
where $\mathcal{M}_{i}\in [0,1]^{|S|\times|O|}$ is the weight matrix of attention initialized as a zero matrix; $a^{o}$ and $a^{s}$ represent the learned attention vector for the object and the state; the $i^{\mathit{th}}$ row $\mathcal{M}_{i}$ and the $j^{\mathit{th}}$ column $\mathcal{M}^{T}_{j}$ in the weight matrix represent the accumulated attention weights for the $i^{\mathit{th}}$ object $o_{i}$ and the $j^{\mathit{th}}$ state $s_{j}$, respectively.

After obtaining the dataset-level weight matrix, we take a few groups of similar objects and their related states from MIT-States as an example to illustrate the effectiveness of SA. We normalize the selected weights along with the axis of the object or state in Figure~\ref{feature_embedding}, showing the learned semantic relations between objects and states. Each row in Figure~\ref{feature_embedding} (a) can be viewed as the state-conditioned object attention; each column in Figure~\ref{feature_embedding} (b) can be viewed as the object-conditioned state attention.

We can observe that SA tends to assign heavy weights to simple primitives that share similar appearances. For example, in the object attention, \textit{peeled} has a high weight score, i.e., feasibility, to fruit, apple, and banana; \textit{dry} relates to sea and river closely. In the state attention, vegetables are highly feasible to show a state of \textit{diced}, \textit{mashed}, \textit{peeled}, \textit{sliced}, or \textit{ripe}; animals and cats show close correlations to \textit{huge}, \textit{large}, and \textit{old}. Obviously, the high feasibility tends to propagate within the same types of objects or states, proving that SA is effective in learning similarity-driven semantics in the datasets. Moreover, no objects have high feasibility to multiple states in the object attention, but \textit{large} and \textit{old} obtain the high feasibility across multiple objects in the state attention. It makes sense because objects may not be described by many types of states while some states (e.g., large) can describe most objects in the real world. \textit{Large} also exhibits a frequent occurrence with \textit{old} having high feasibility to the same object, suggesting that many \textit{large} objects may be also described by \textit{old}. It is common, especially when describing animals, e.g., larger cats usually equal older cats.

\begin{table}[t]
    \centering
    \caption{\textbf{Left two columns}: Examples of Bottom-3 feasible compositions in SA on MIT-States and C-GQA datasets. \textbf{Right two columns}: Examples of Top-3 feasible seen compositions relating to the infeasible objects and states in the left two columns.}
    \resizebox{.95\linewidth}{!}{
   \begin{tabular}{cc|cc}
   \midrule
\multicolumn{4}{c}{MIT-States}                                                   \\
State       & Bottom-3 Feasible Objects      & Infeasible Object    & Top-3 Seen States     \\
\midrule
bent        & flame,vacuum,drum              & flame     & molten,brushed        \\
lightweight & vacuum,flame,laptop            & laptop    & black,open,silver     \\
\midrule
Object      & Bottom-3 Feasible States       & Infeasible State     & Top-3  Seen Objects   \\
\midrule
book        & standing,dull,short            & standing  & tower                 \\
bucket      & blunt,mashed,standing          & mashed    & bean,vegetable,banana \\
\midrule
\multicolumn{4}{c}{C-GQA}                                                        \\
State       & Bottom-3 Feasible Objects      & Infeasible Object    & Top-3 Seen States     \\
\midrule
artificial  & shield,charger,courtyard       & shield    & glass,protective      \\
misty       & briefcase,deer,antenna         & briefcase & blue,black,clear      \\
\midrule
Object      & Bottom-3 Feasible States       & Infeasible State     & Top-3  Seen Objects   \\
\midrule
cauliflower & feathered,rustic,winding       & feathered & wing                  \\
eagle       & connected,discolored,miniature & connected & chain,cord           \\
\midrule
\end{tabular}}
    
    \label{semantics_attention_example_bottom}
\end{table}

\begin{figure*}[t]
    \centering 
\hspace{-1mm}
\begin{subfigure}{0.24\textwidth}
  \includegraphics[width=\textwidth]{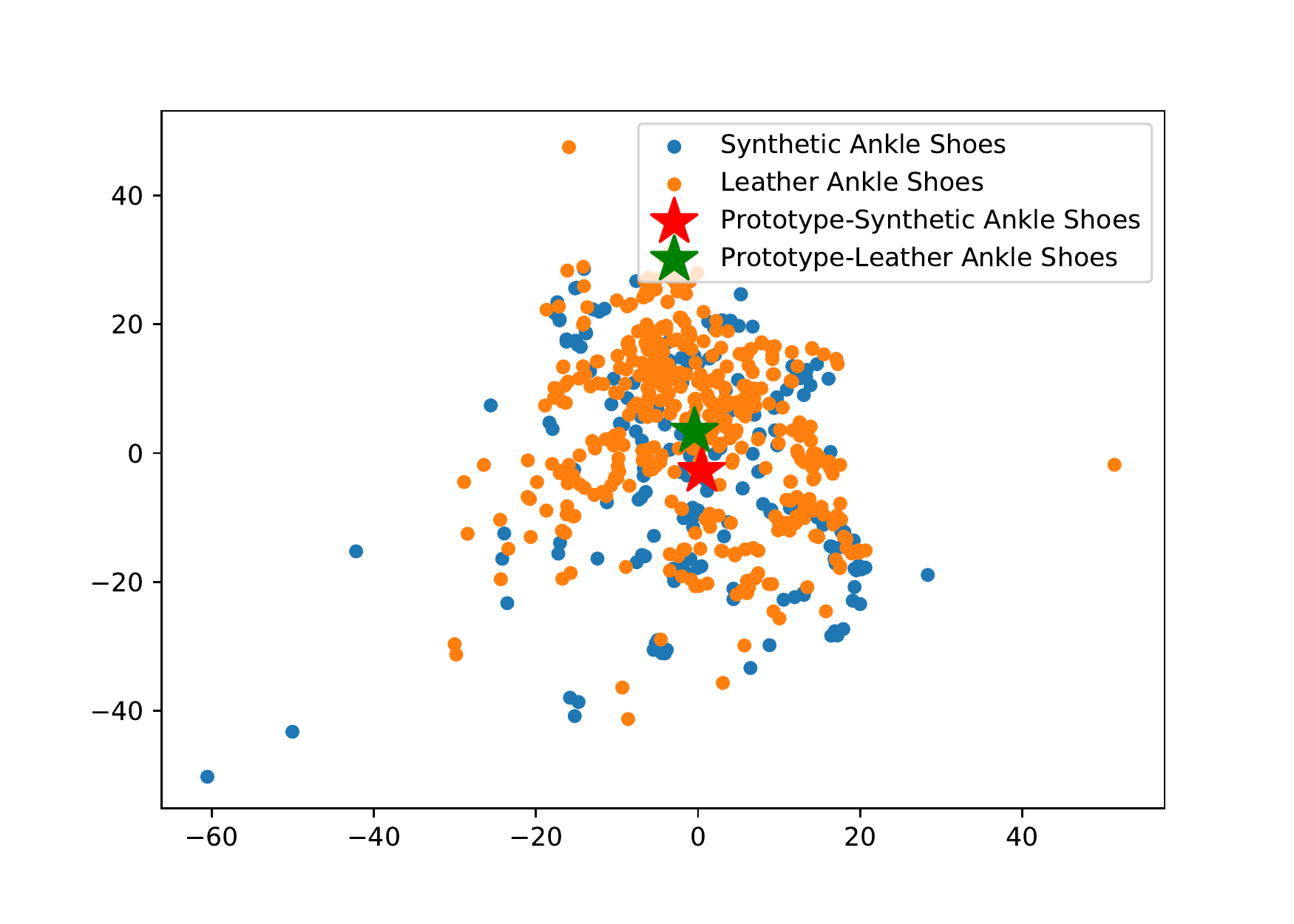}
    \centering
    \caption{Original Object.}
\end{subfigure}
\begin{subfigure}{0.24\textwidth}
  \includegraphics[width=\textwidth]{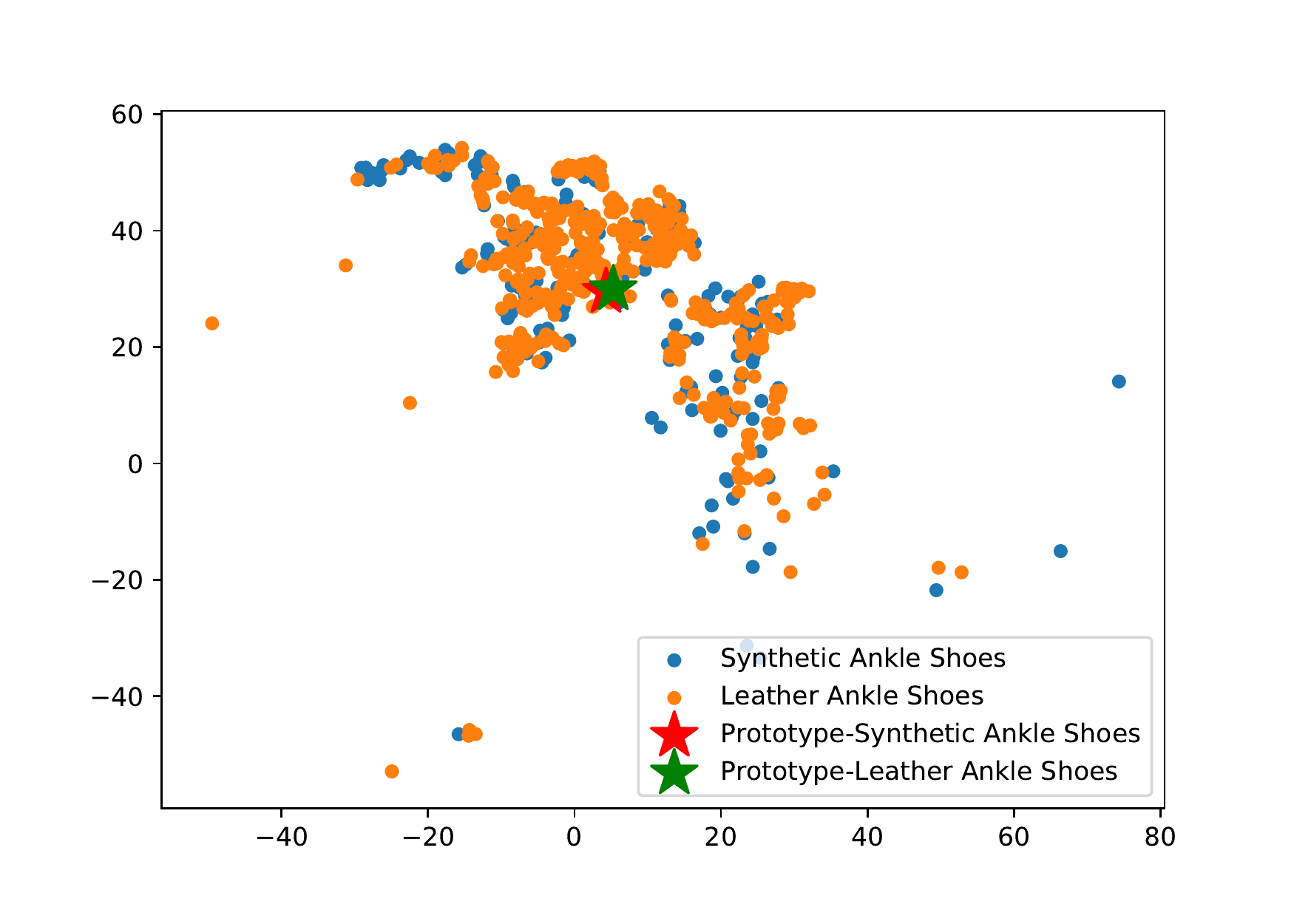}
    \centering
    \caption{Disentangled Object.}
\end{subfigure}
\begin{subfigure}{0.24\textwidth}
  \includegraphics[width=\textwidth]{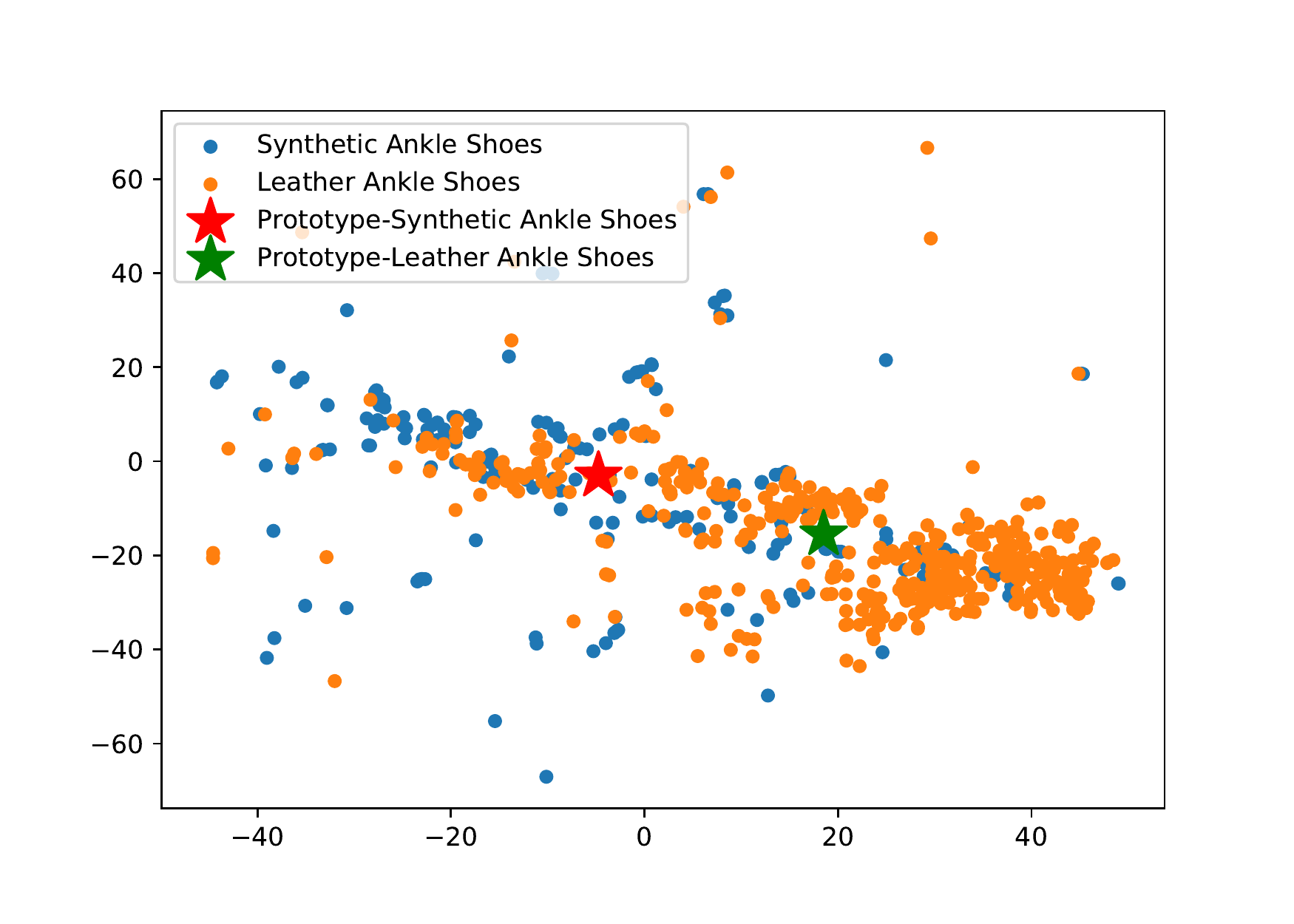}
    \centering
    \caption{Original State.}
\end{subfigure}
\begin{subfigure}{0.24\textwidth}
  \includegraphics[width=\textwidth]{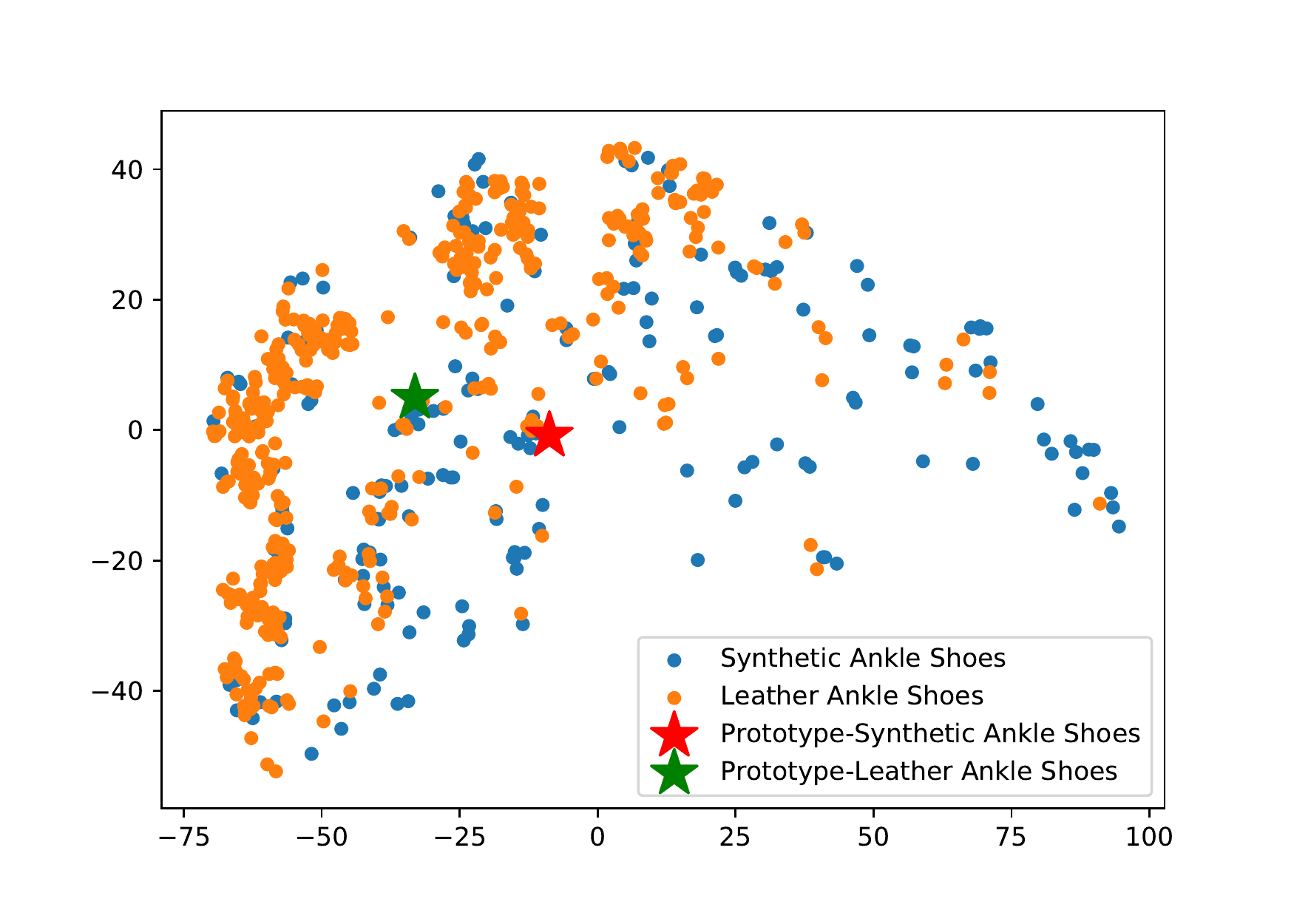}
    \centering
    \caption{Disentangled State.}
\end{subfigure}
\caption{Original feature embedding and disentangled feature embedding for \textit{synthetic ankle shoes} and \textit{leather ankle shoes}. Figures (a-b) plot the original and disentangled object embeddings for \textit{ankle shoes}, respectively. Figures (c-d) plot the original and disentangled state embeddings for \textit{synthetic} and \textit{leather}, respectively.}
\label{same_object_dif_state_embedding}
\end{figure*}

\begin{figure*}[t]
    \centering 
\hspace{-1mm}
\begin{subfigure}{0.24\textwidth}
  \includegraphics[width=\textwidth]{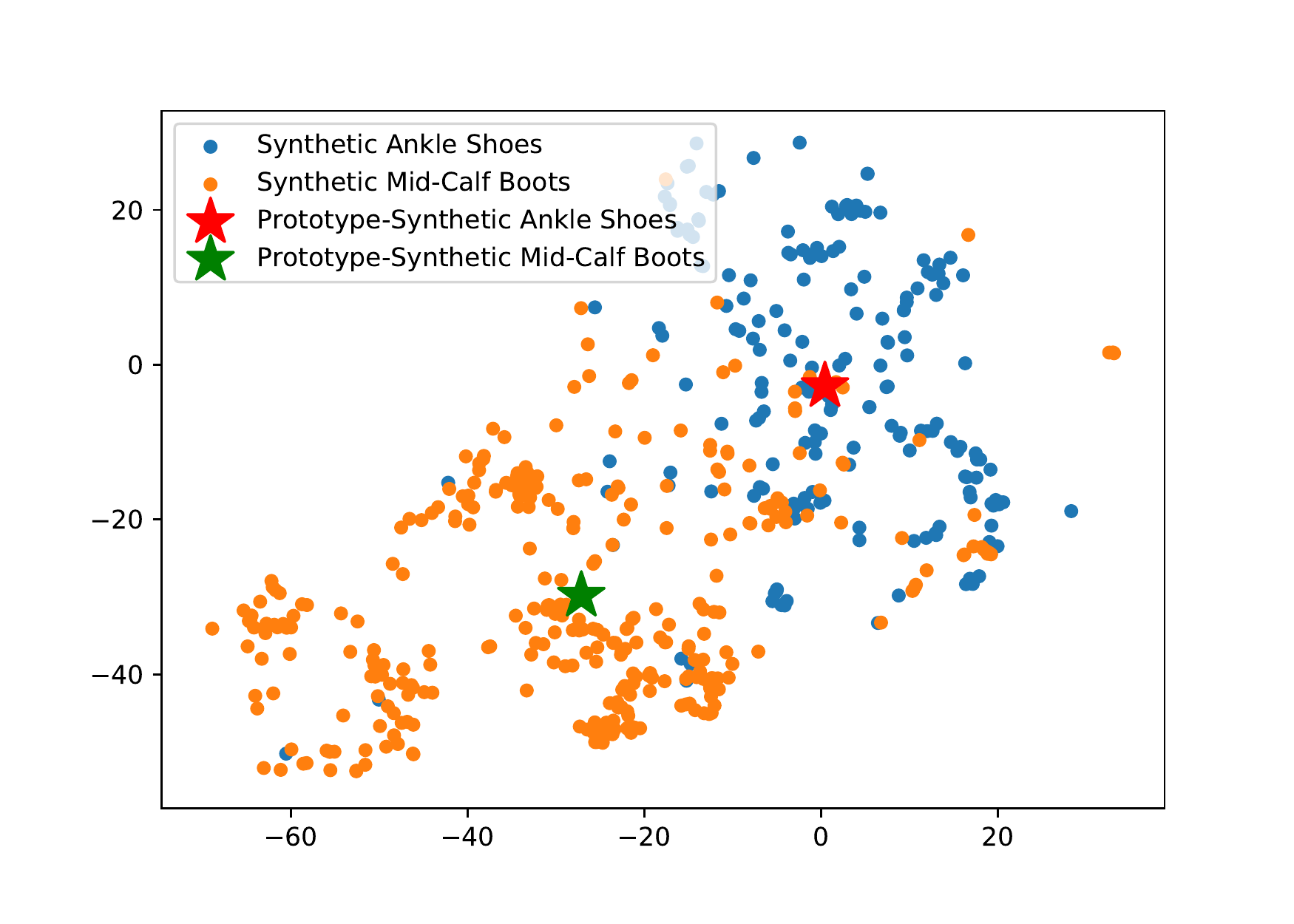}
    \centering
    \caption{Original Object.}
\end{subfigure}
\begin{subfigure}{0.24\textwidth}
  \includegraphics[width=\textwidth]{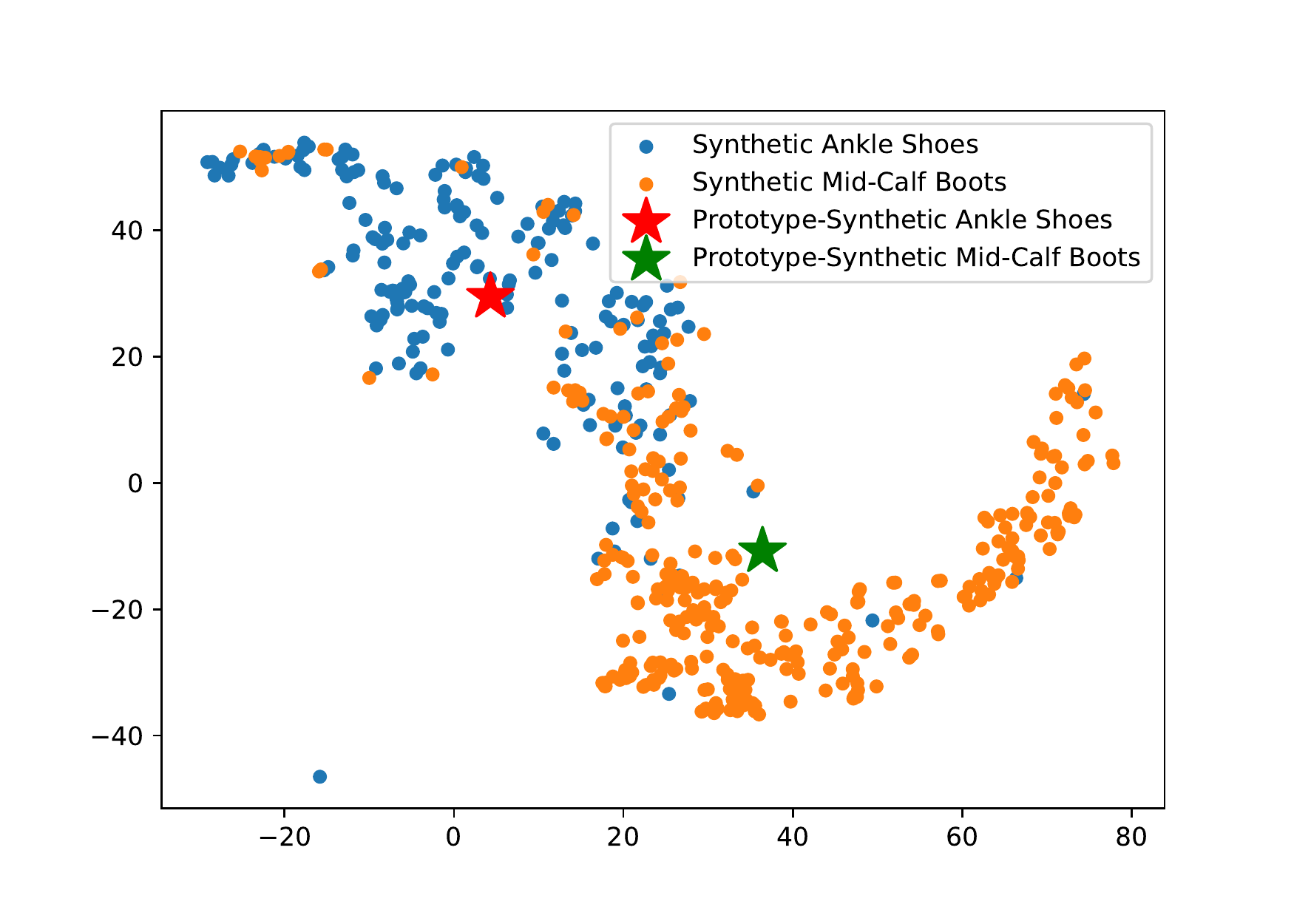}
    \centering
    \caption{Disentangled Object.}
\end{subfigure}
\begin{subfigure}{0.24\textwidth}
  \includegraphics[width=\textwidth]{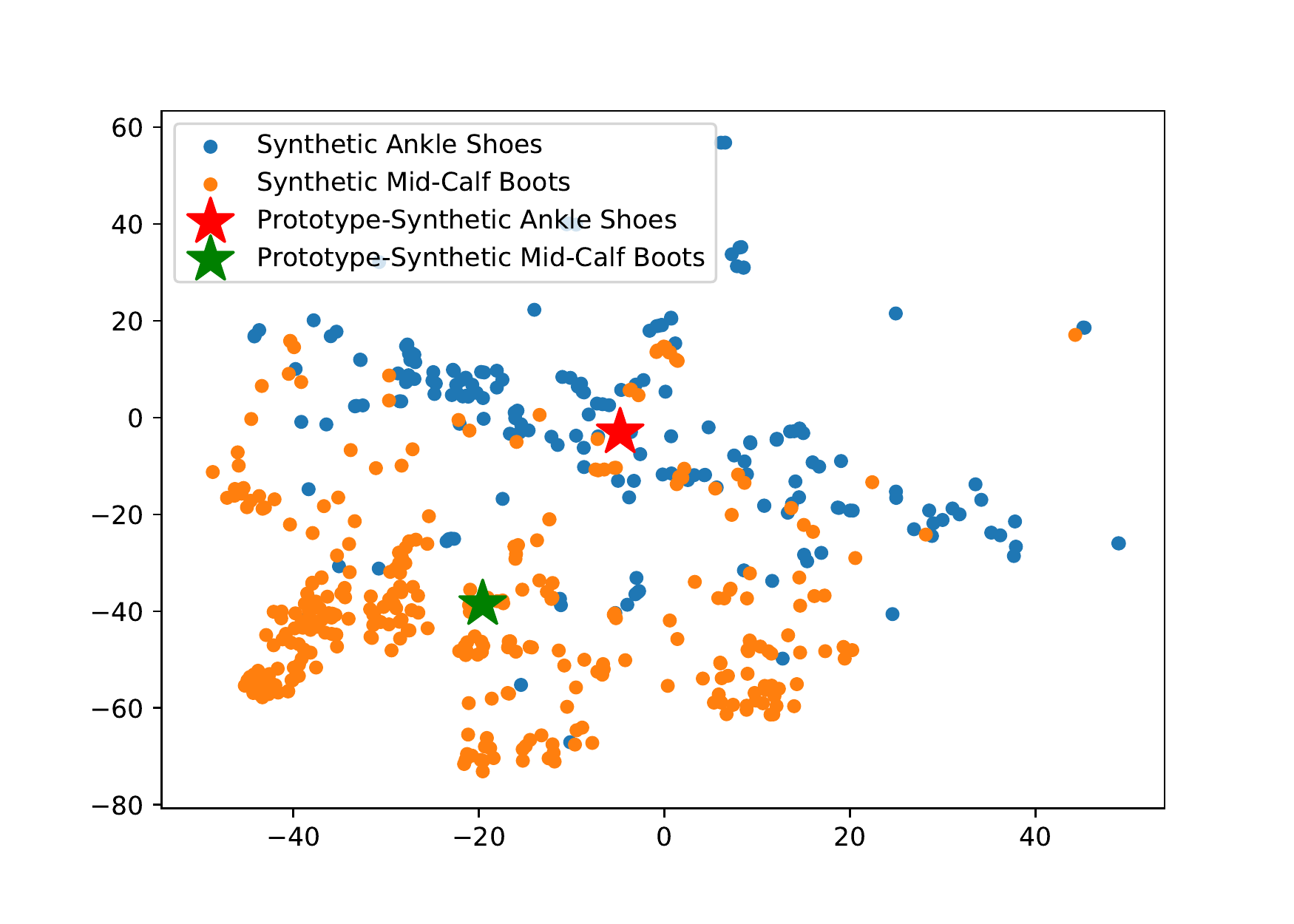}
    \centering
    \caption{Original State.}
\end{subfigure}
\begin{subfigure}{0.24\textwidth}
  \includegraphics[width=\textwidth]{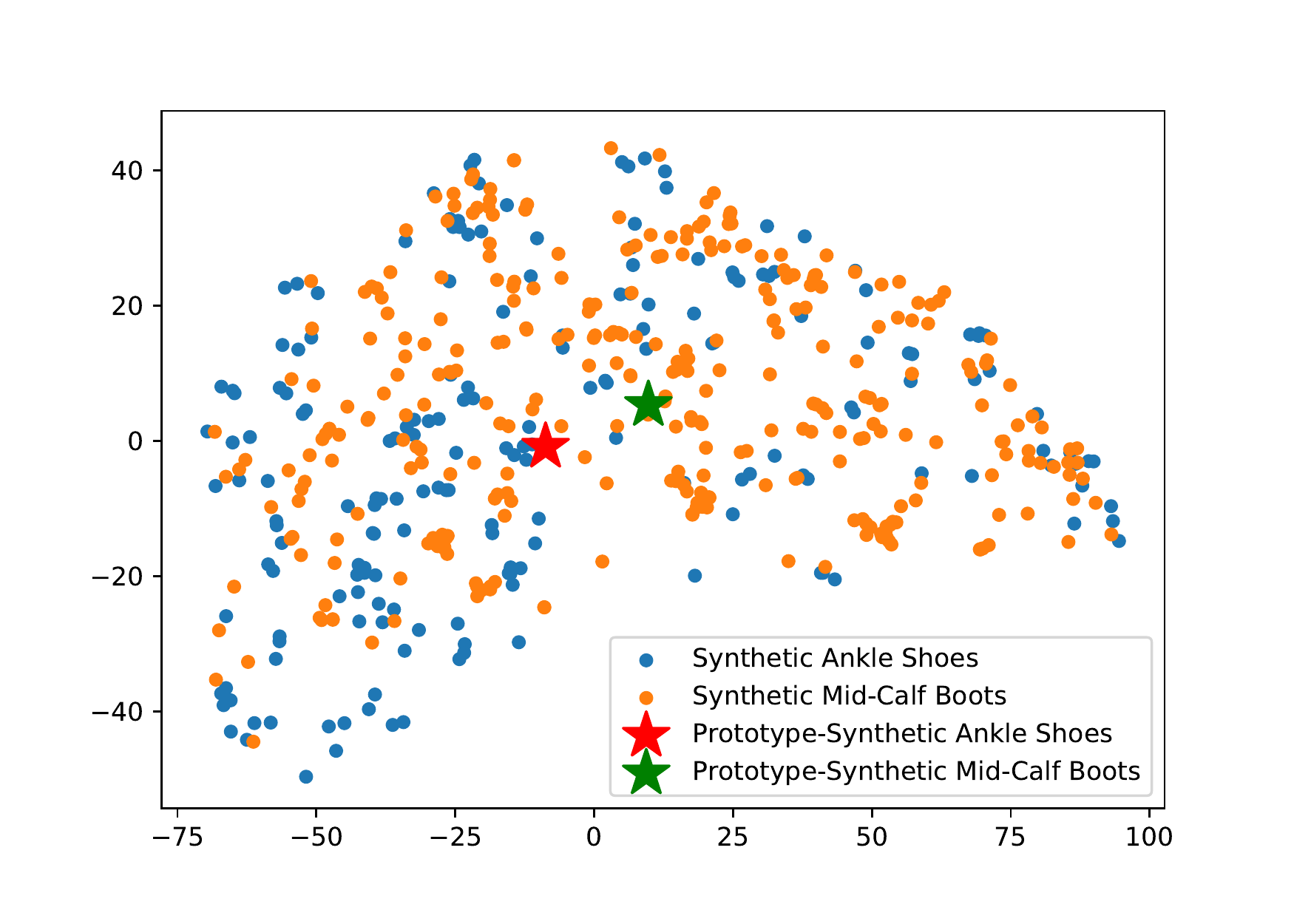}
    \centering
    \caption{Disentangled State.}
\end{subfigure}
\caption{Original feature embedding and disentangled feature embedding for \textit{synthetic ankle shoes} and \textit{synthetic mid-calf boots}. Figures (a-b) plot the original and disentangled object embeddings for \textit{ankle shoes} and \textit{mid-calf boots}, respectively. Figures (c-d) plot the original and disentangled state embeddings for \textit{synthetic}, respectively.}
\label{same_state_dif_obj_embedding}
\end{figure*}
\subsubsection{Feasible Compositions in Semantic Attention}

In this section, we demonstrate the effectiveness of SA by showing the most/least feasible compositions based on the frequency. We argue that the more frequently a composition is assigned the highest/lowest weight in the attention map, the more/less feasible it is considered by the attention mechanism. Therefore, we count the frequency and obtain the most frequent state-object pairs, representing the most/least feasible compositions, i.e., the most frequent compositions with the largest/least weight in $p_{f}(s|o)$ and $p_{f}(o|s)$, for each state and object. We take MIT-States and C-GQA datasets as examples, showing the Top/Bottom-3 feasible compositions (i.e., Top/Bottom-3 most feasible compositions with the highest/lowest attention weight) in Table~\ref{semantics_attention_example} and Table \ref{semantics_attention_example_bottom}, respectively.

In Table~\ref{semantics_attention_example}, we exhibit the Top-3 feasible compositions conditioned on objects and states. We show the most feasible compositions in two different settings: the Open-World space (OW) and the Unseen compositional (UC) space. OW may contain compositions from seen compositions while UC only contains unseen compositions. Thus, we can know whether SA can infer feasible unseen compositions by only learning the existing semantics of datasets. We can observe that SA effectively finds the GT unseen compositions and views them as the one of most feasible compositions. For example, the GT unseen composition, i.e., straight blade, is viewed as a Top-3 feasible composition for \textit{straight} under both OW and UC settings. Blue tail, hairy tail, and long tail are the most feasible compositions for a tail while only \textit{long tail} is seen during training. These findings prove that SA can predict GT unseen compositions by learning the seen semantics in datasets.

In Table~\ref{semantics_attention_example_bottom}, we can observe that the most infeasible compositions have a low probability of existing in the real world, e.g., bent flame and feathered cauliflower. There are two possible reasons that SA considers these compositions infeasible. 1) The related images of simple primitives are too distinct from other images. For example, seen images that relate to flame are \textit{molten flame} and \textit{brushed flame}. These distinct images share little similarity with other images. 2) Few related seen compositions exist in datasets. For example, \textit{feathered} only relates to \textit{wing} during training. SA propagates little feasibility based on \textit{feathered} and thus \textit{feathered} has no close correlations to other simple primitives. The infeasible compositions suggest that our method is capable of finding those distinct simple primitives, and then assigning low feasibility to their unrelated concepts.

\subsubsection{Unbiased Features of Knowledge Disentanglement}

In this section, we project features to two-dimensional embeddings based on t-SNE~\cite{van2008visualizing}. Then, we visualize the projected embeddings of original features and disentangled features provided by KD, showing whether the distributional loss $\mathcal{L}_{kd}$ can supervise KD to disentangle features. In Figure~\ref{same_object_dif_state_embedding} and Figure~\ref{same_state_dif_obj_embedding}, we view the mean embedding of simple primitives as a prototype to show the center point of embedding distributions, and we exhibit two pairs of compositions on UT-Zappos. Figures~\ref{same_object_dif_state_embedding} (a-d) visualize the composition pair \textit{synthetic ankle shoes} and \textit{leather ankle shoes}, which share the same object (i.e., ankle shoes) but different states (i.e., synthetic and leather). In Figures~\ref{same_object_dif_state_embedding} (a-b), KD can reduce the distance of the object prototype of ankle shoes significantly. Thus, the embedding space of ankle shoes is more concentrated after disentanglement. Figures~\ref{same_object_dif_state_embedding} (c-d) share the similar prototype distance, but nodes are more dispersed after disentanglement. In Figures~\ref{same_state_dif_obj_embedding} (a-d), we plot compositions that share the same state but different objects, i.e., synthetic ankle shoes and synthetic mid-calf boots. From Figures~\ref{same_state_dif_obj_embedding} (a-b), KD can increase the distance of object prototype of ankle shoes and mid-calf boots. The node distributions of ankle shoes and mid-calf boots are more concentrated around their prototype after disentanglement. Figures~\ref{same_state_dif_obj_embedding} (c-d) show that the distance between the state prototypes of \textit{synthetic} is largely reduced, which means that KD can effectively extract the state information from different objects.
In conclusion, KD supervised by $\mathcal{L}_{kd}$ can effectively disentangle states/objects to learn unbiased feature representations. The unbiased feature representations of the same state/object in different compositions are clustered; the feature distributions of different objects/states are dispersed. Thus, KD can learn more disentangled and unbiased features for classification than the original features, which may ease the biased predictions caused by contextuality. 

\begin{figure*}[t]
    \centering 
\hspace{-1mm}
\includegraphics[width=\textwidth]{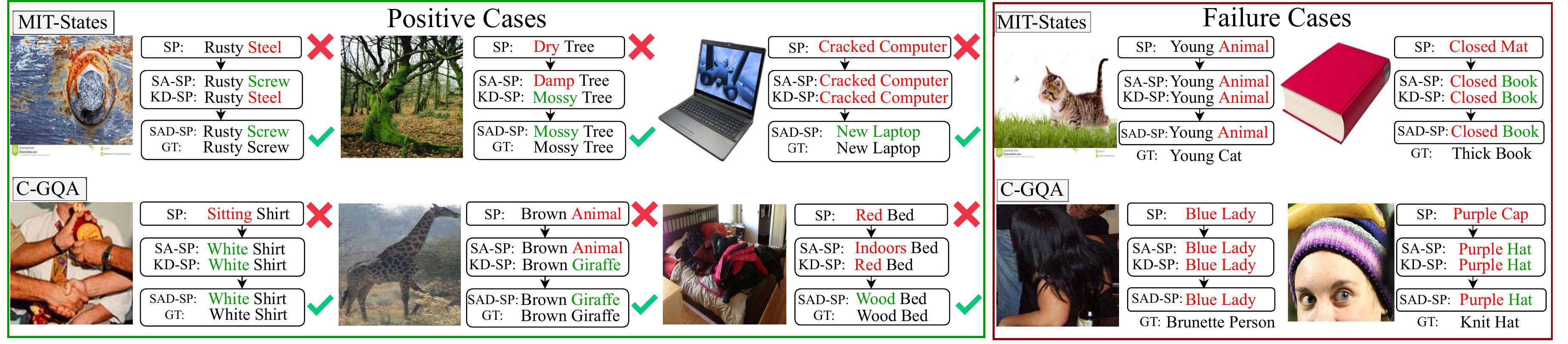}
\caption{Case study on MIT-States and C-GQA.}
\label{decision_example}
\end{figure*}

\subsubsection{Case Study}\label{case_study}
We conduct a case study to illustrate how SAD-SP revises the predictions of SP and the limitations of our model in Figure ~\ref{decision_example}. In the positive cases, SA can calibrate \textit{steel} to \textit{screw} and \textit{sitting} to \textit{white}, which is more consistent with the semantic relations in the dataset. KD enables SAD-SP to identify the moss on the tree and to refine animals to giraffes, which indicates the improvements in the ability to recognize simple primitives in our model. Moreover, the synergy of SA and KD can fix some misclassifications in SP, SA-SP, and KD-SP, such as modifying a cracked computer into a new laptop and letting the model focus on the wooden material of a bed. However, when there are multiple simple primitives existing in the image, our model tends to make wrong predictions. For example, our model fails to refine an animal as a cat when the cat is hidden in the grass, and an insect flies around it. It also views the blue color from the background as the state of the main object mistakenly. These mistakes indicate that our model sometimes fails to locate the main object or state in the image. The possible reason is that the model and annotator view different things as the main object in the image. While SAD-SP can partially ease this issue, e.g., SA improves \textit{red} into \textit{indoor}, or SAD-SP enhances \textit{red} into \textit{wood}, it still misclassifies a knit hat as a purple hat, a person with long hair as a lady, and a thick book as a closed book. These predictions may not be mistakes strictly, but it shows the deficiency of our model in learning the human preference for label annotations, e.g., SAD-SP is not aware that color may have the lowest priority when describing objects.

\subsubsection{Summary and Discussion of Dependence}
\textbf{Summary of experiments.} Compared with the SOTA methods in OW-CZSL, SAD-SP shows competitive or better performance using fine-tuned and fixed backbone networks. We propose two new parallel networks, SA and KD, which can enhance SP by learning feasibility- and contextuality-dependence. We demonstrate the effectiveness of SA and KD from multiple aspects: 1) Ablation studies of each module and branch demonstrate the effectiveness of the proposed modules in the open-world generalized compositional zero-shot learning. 2) The studies of the learned weight distribution and frequency show that the feasibility of SA learning makes use of visual similarity effectively, accurately inferring the unseen compositions based on the existing semantic relations in datasets; the visualization of feature representations proves that the proposed distributional loss is able to supervise KD to learn unbiased disentangled features, thereby obtaining a better feature distribution. 3) Positive and failure case studies reveal that SA and KD can effectively revise the wrong predictions of SP and partially learn the priority in the manual annotation.

\textbf{Discussion of datasets.} From the experiments, we can observe that the advances of SAD-SP on datasets are owing to different modules. Since we use the same setting for training, e.g., network structures, learning rates, and hyper-parameters, we consider the learning abilities of our models are similar. We argue that challenges constraining the progress of SOTA methods are different in datasets. UT-Zappos is a distinct dataset that only contains shoes and common materials. Since no strict limitations of materials are posed on making shoes, feasibility-dependence provides little information for improving the predictions. On the contrary, contextuality-dependence is essential to learn, which plays an important role in discriminating entangled shoe types and materials. This is why KD contributes the most to SAD-SP on UT-Zappos in identifying footwear components. Different from UT-Zappos, MIT-States and C-GQA datasets consist of objects from a wide range, e.g., animals and buildings. Due to the diverse objects, feasibility-dependence is informative and essential to the performance improvement of SAD-SP. Meanwhile, the wider range of objects and related states makes it more difficult to recognize simple primitives, especially states. On MIT-States, our model is still capable of learning the unbiased states and objects beneficial for predictions. However, on C-GQA, the current network structure may not be capable of handling the extremely large scope of simple primitives ($\sim$4 times the states and $\sim$2 times the objects of MIT-States) well, leading to SAD-SP only achieving the best performance after disabling some branches. We speculate that datasets with more diverse simple primitives require stronger learning capabilities of feasibility- and contextuality-dependence.

\textbf{Limitations.} Though branches in SA and KD can achieve robust results under a unified setting, the fixed hyper-parameters result in SAD-SP not being able to dynamically balance feasibility and context dependencies. For example, SAD-SP cannot disable invalid branches dynamically may result in a sub-optimal output. In addition, while SA and KD can alleviate some biased predictions in SP, such as refining a vague concept into an explicit class, and denoising the irrelevant visual information in the background, SAD-SP may still misunderstand the appropriate simple primitives or the main object in the images.
\begin{table}[t]
\centering
\caption{Accuracy of state and object predictions when achieving best HM and U on the UT-Zappos testing set. HM state and HM object represent the accuracy of the overall state and object predictions (including seen and unseen classes) when the model achieves the best HM scores. Unseen state and unseen object represent the accuracy of the unseen state and object predictions when the model achieves the best U scores.}
\resizebox{.91\linewidth}{!}{
\begin{tabular}{|c|c|c|c|c|}
\hline
       & HM State & HM Object & Unseen State & Unseen Object \\ \hline
SP     & 54.04   & 73.64    & 42.10       & 71.82        \\ \hline
KD-SP  & 54.49   & 74.29    & 40.94        & 74.33        \\ \hline
SA-SP  & 55.28   & 73.54    & 42.75       & 70.87        \\ \hline
SAD-SP & 54.63    & 73.75     & 42.21        & 72.24         \\ \hline
\end{tabular}}
\label{independent_state_object_pred}
\end{table}

\begin{figure*}[t]
    \centering
    \subfloat[]{\includegraphics[width=0.33\textwidth]{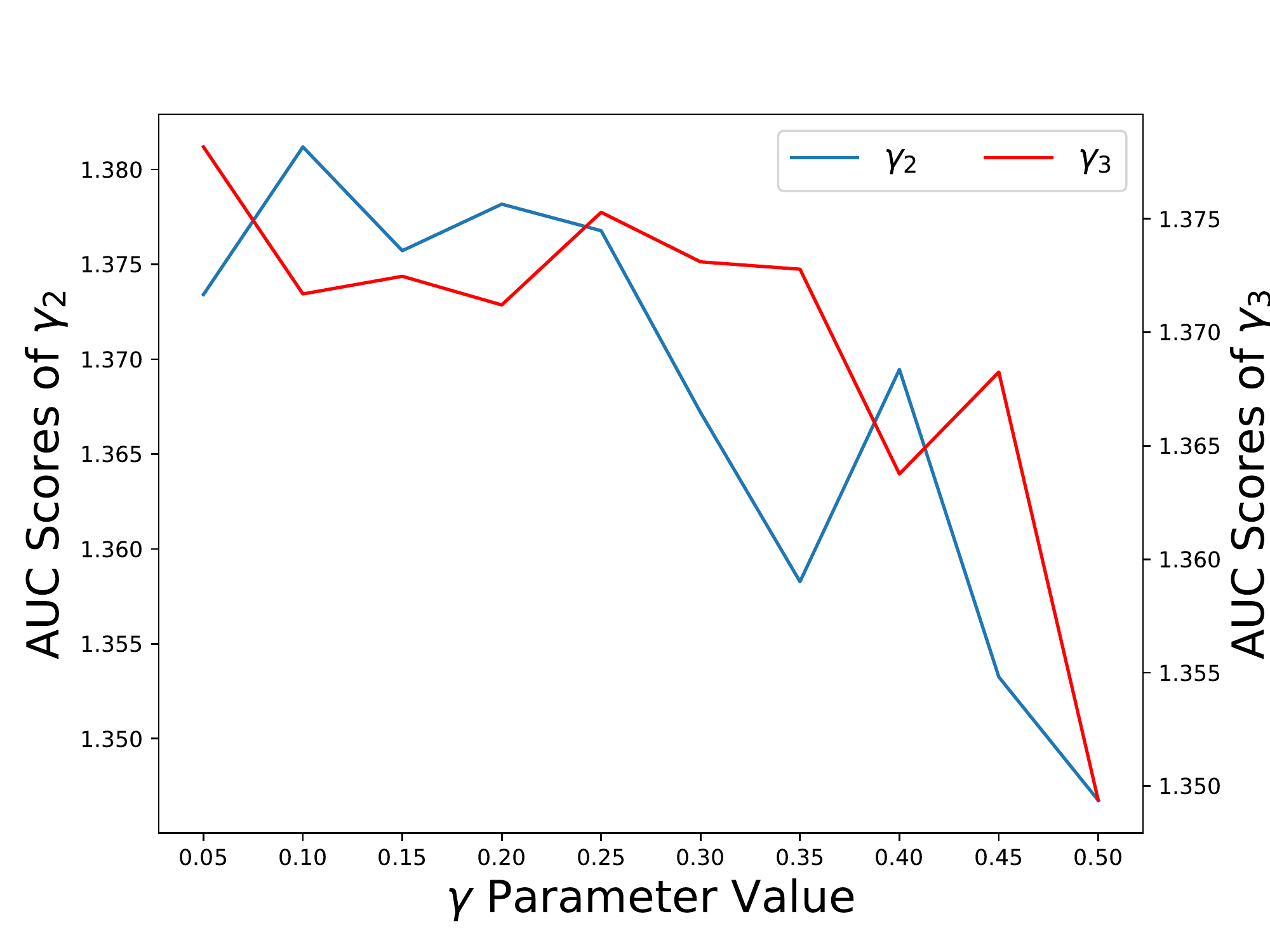}}\hfil 
    \subfloat[]{\includegraphics[width=0.33\textwidth]{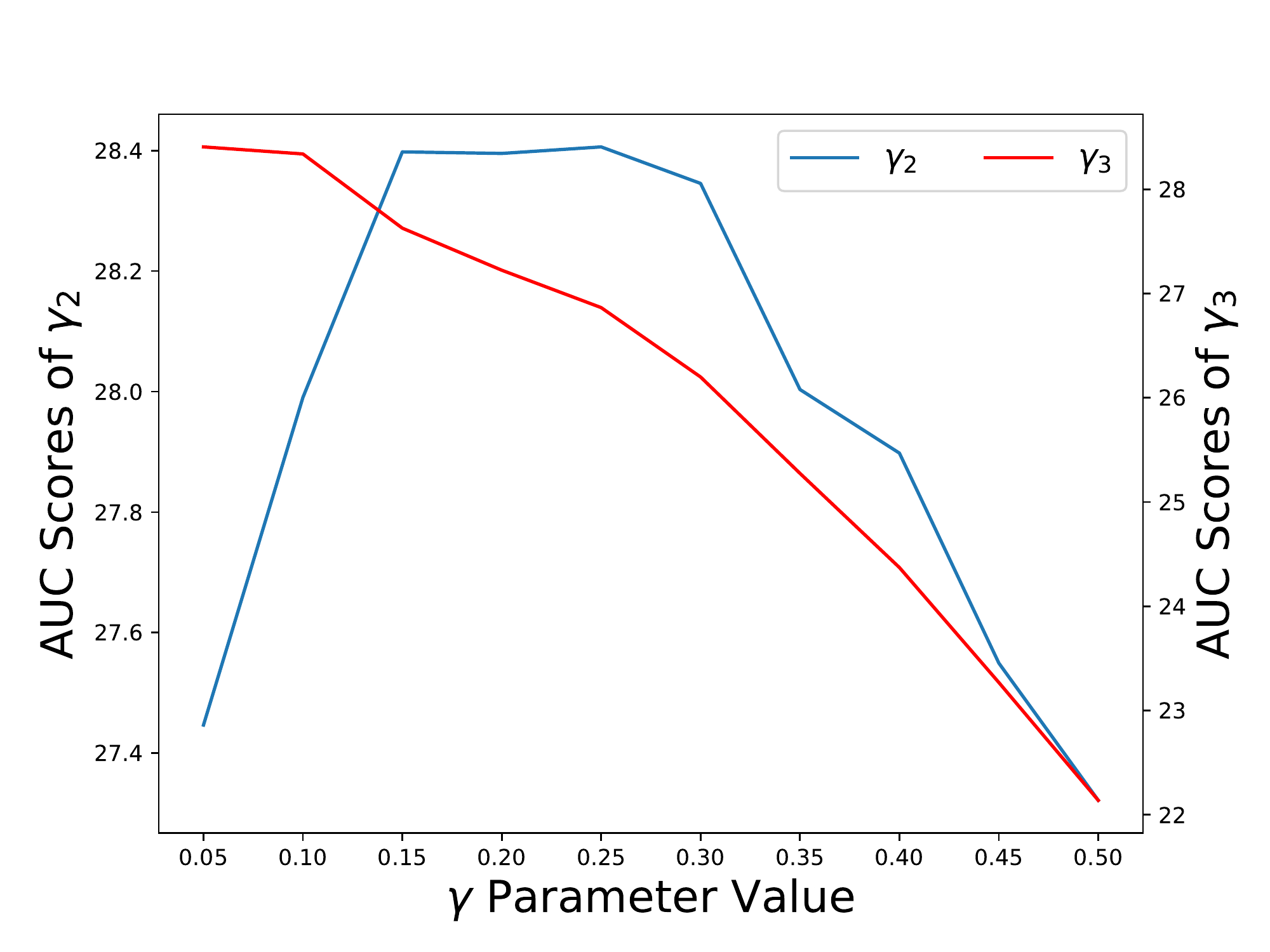}}\hfil 
    \subfloat[]{\includegraphics[width=0.33\textwidth]{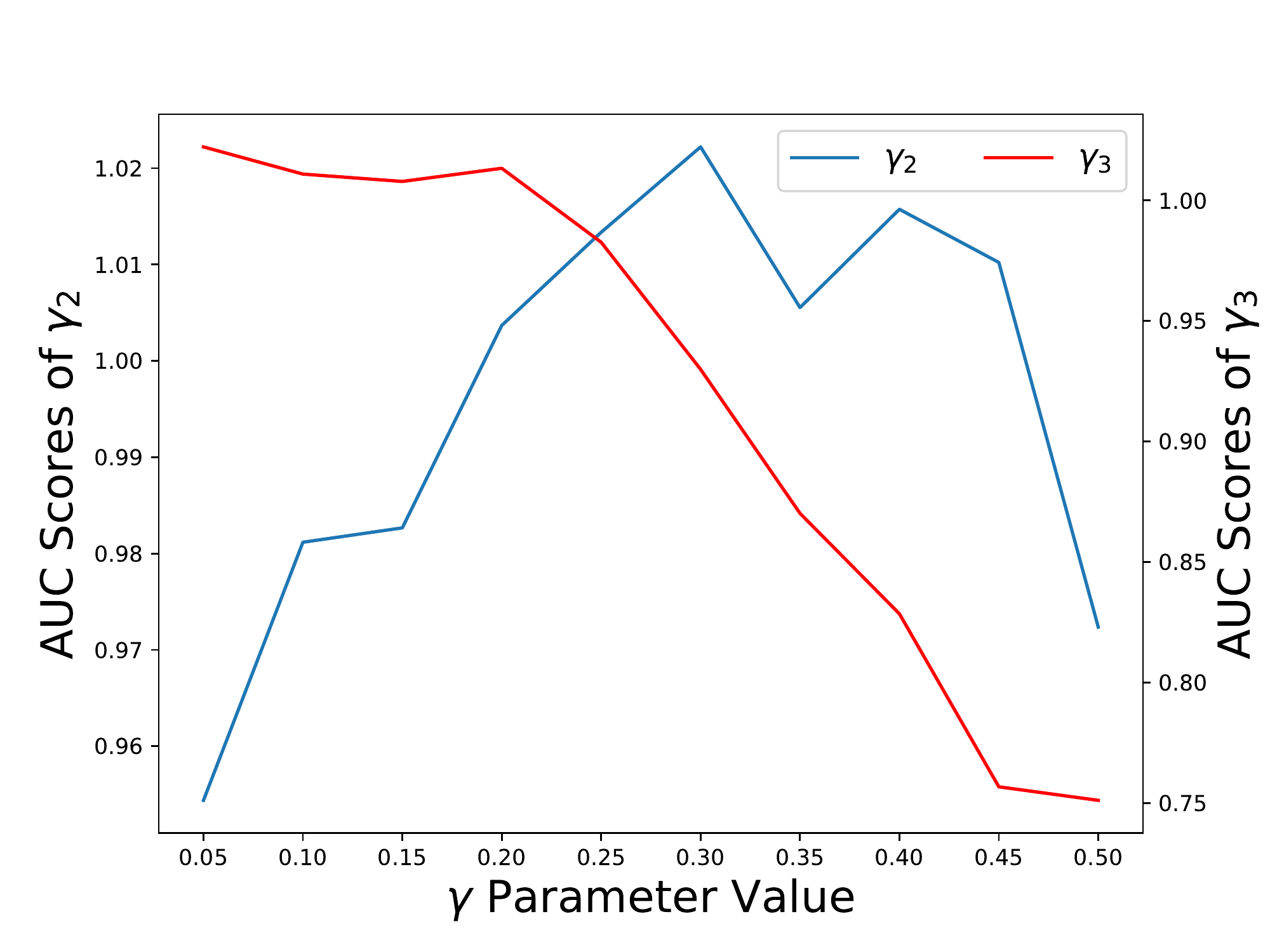}}\hfil 
\caption{Best AUC scores of SAD-SP on three datasets with $\gamma_{2}$ and $\gamma_{3}$ ranging from 0.05 to 0.5: (a) MIT-States, (b) UT-Zappos, and (c) C-GQA.}
\label{supp_snet_analysis}
\end{figure*}

\subsection{Analysis of Object and State Predictions} 
In Table~\ref{independent_state_object_pred}, we take the UT-Zappos dataset as an example to exhibit the detailed accuracy of state and object predictions when achieving the best HM and U scores. From the table, we can observe that states in UT-Zappos are much more difficult to recognize than objects. Moreover, SA shows better performance in improving state accuracy, and KD enhances object recognition. SAD-SP may not obtain the highest state or object accuracy but still can achieve the best AUC scores, which means that SAD-SP can balance the learning biases of SA and KD to obtain the optimal composition-wise accuracy. Improving the accuracy of state or object predictions independently may not lead to an increase in composition-wise predictions. It is more important to learn the dependence between simple primitives, letting the simple primitives in the compositions be predicted correctly at the same time.

\subsection{Hyper-parameter Study} 
This section studies the influences of hyper-parameter $\gamma$ on the model performance. We separately vary the ratios of feasibility- and contextuality-dependence in predictions, i.e., $\gamma_{2}$ and $\gamma_{3}$, ranging from 0.05 to 0.5 to investigate the effects of $\gamma$. In Figure~\ref{supp_snet_analysis}, we fix the values of $\gamma_{2}$ and $\gamma_{3}$, showing the corresponding values on the x-axis. We vary the unfixed hyper-parameters to search the highest AUC score with the fixed $\gamma_{2}$ or $\gamma_{3}$, exhibiting the results on the left and right sides of the y-axis, respectively. The model achieves the best performance when $\gamma_{2}$ is in the range of 0.1 to 0.25 and $\gamma_{3}$ is in the range of 0.05 to 0.2. When $\gamma_{2}$ and $\gamma_{3}$ are larger than 0.35, the model performance will significantly drop, especially when $\gamma_{2}=0.5$ or $\gamma_{3}=0.5$. Combined with the patterns in the analysis of state/object predictions, this phenomenon suggests that large $\gamma_{2}$ or $\gamma_{3}$ will lead to biased predictions towards objects or states, which will reduce the generalization ability to predict unseen compositions in OW-CZSL. Therefore, we select $\gamma_{2}=0.25$ and $\gamma_{3}=0.05$ as the balanced parameters.





\section{Conclusion}
In this work, we propose a Semantic Attention and knowledge Disentanglement guided Simple Primitives (SAD-SP) to tackle the deficiency of feasibility and contextuality in simple primitives under the open-world compositional zero-shot learning setting. We design semantic attention, which learns to infer the feasibility relations in datasets driven by visual similarity, to provide auxiliary classification information for predictions. We also propose a generative knowledge disentanglement module, which conducts knowledge disentanglement supervised by a distributional loss, to learn unbiased feature representations specific to states and objects, easing the biased predictions caused by contextuality. Via experiments, we discuss the underlying reasons for our improvement on different datasets, providing some insightful analysis for the community. We conclude some potential challenges and limitations in OW-CZSL, e.g., the limited semantic relations between shoes and materials on UT-Zappos, the difficult state recognition on MIT-States and C-GQA, and the difficulty in learning the annotation priority of humans. In the future, we plan to implement dynamic $\gamma$ selections and enhance the network structure of the backbone~\cite{jin2021channel,zhang2022deep}, which may fix the learning deficiency and balance the learning tendencies dynamically. In this way, our model, SAD-SP, may be capable of addressing the challenges found in our experiments, especially when tackling datasets with large numbers of simple primitives.

%
%


%





\ifCLASSOPTIONcaptionsoff
  \newpage
\fi
\bibliographystyle{IEEEtran}
\bibliography{bio}

\begin{thebibliography}{10}
\providecommand{\url}[1]{#1}
\csname url@samestyle\endcsname
\providecommand{\newblock}{\relax}
\providecommand{\bibinfo}[2]{#2}
\providecommand{\BIBentrySTDinterwordspacing}{\spaceskip=0pt\relax}
\providecommand{\BIBentryALTinterwordstretchfactor}{4}
\providecommand{\BIBentryALTinterwordspacing}{\spaceskip=\fontdimen2\font plus
\BIBentryALTinterwordstretchfactor\fontdimen3\font minus
  \fontdimen4\font\relax}
\providecommand{\BIBforeignlanguage}[2]{{%
\expandafter\ifx\csname l@#1\endcsname\relax
\typeout{** WARNING: IEEEtran.bst: No hyphenation pattern has been}%
\typeout{** loaded for the language `#1'. Using the pattern for}%
\typeout{** the default language instead.}%
\else
\language=\csname l@#1\endcsname
\fi
#2}}
\providecommand{\BIBdecl}{\relax}
\BIBdecl

\bibitem{zhang2017range}
X.~Zhang, Z.~Fang, Y.~Wen, Z.~Li, and Y.~Qiao, ``Range loss for deep face
  recognition with long-tailed training data,'' in \emph{Proceedings of the
  IEEE International Conference on Computer Vision}, 2017, pp. 5409--5418.

\bibitem{zhao2021hierarchical}
H.~Zhao, S.~Guo, and Y.~Lin, ``Hierarchical classification of data with
  long-tailed distributions via global and local granulation,''
  \emph{Information Sciences}, vol. 581, pp. 536--552, 2021.

\bibitem{socher2013zero}
R.~Socher, M.~Ganjoo, C.~D. Manning, and A.~Ng, ``Zero-shot learning through
  cross-modal transfer,'' \emph{Advances in neural information processing
  systems}, vol.~26, 2013.

\bibitem{zhang2021bag}
Y.~Zhang, X.-S. Wei, B.~Zhou, and J.~Wu, ``Bag of tricks for long-tailed visual
  recognition with deep convolutional neural networks,'' in \emph{Proceedings
  of the AAAI conference on artificial intelligence}, vol.~35, no.~4, 2021, pp.
  3447--3455.

\bibitem{samuel2021generalized}
D.~Samuel, Y.~Atzmon, and G.~Chechik, ``From generalized zero-shot learning to
  long-tail with class descriptors,'' in \emph{Proceedings of the IEEE/CVF
  Winter Conference on Applications of Computer Vision}, 2021, pp. 286--295.

\bibitem{yan2022semantics}
C.~Yan, X.~Chang, M.~Luo, H.~Liu, X.~Zhang, and Q.~Zheng, ``Semantics-guided
  contrastive network for zero-shot object detection,'' \emph{IEEE Transactions
  on Pattern Analysis and Machine Intelligence}, 2022.

\bibitem{pourpanah2022review}
F.~Pourpanah, M.~Abdar, Y.~Luo, X.~Zhou, R.~Wang, C.~P. Lim, X.-Z. Wang, and
  Q.~J. Wu, ``A review of generalized zero-shot learning methods,'' \emph{IEEE
  transactions on pattern analysis and machine intelligence}, 2022.

\bibitem{romera2015embarrassingly}
B.~Romera-Paredes and P.~Torr, ``An embarrassingly simple approach to zero-shot
  learning,'' in \emph{International Conference on Machine Learning}, 2015, pp.
  2152--2161.

\bibitem{wang2019survey}
W.~Wang, V.~W. Zheng, H.~Yu, and C.~Miao, ``A survey of zero-shot learning:
  Settings, methods, and applications,'' \emph{ACM Transactions on Intelligent
  Systems and Technology (TIST)}, vol.~10, no.~2, pp. 1--37, 2019.

\bibitem{wang2020generalizing}
Y.~Wang, Q.~Yao, J.~T. Kwok, and L.~M. Ni, ``Generalizing from a few examples:
  A survey on few-shot learning,'' \emph{ACM computing surveys (csur)},
  vol.~53, no.~3, pp. 1--34, 2020.

\bibitem{hospedales2021meta}
T.~Hospedales, A.~Antoniou, P.~Micaelli, and A.~Storkey, ``Meta-learning in
  neural networks: A survey,'' \emph{IEEE transactions on pattern analysis and
  machine intelligence}, vol.~44, no.~9, pp. 5149--5169, 2021.

\bibitem{ruis2021independent}
F.~Ruis, G.~Burghouts, and D.~Bucur, ``Independent prototype propagation for
  zero-shot compositionality,'' \emph{Advances in Neural Information Processing
  Systems}, vol.~34, pp. 10\,641--10\,653, 2021.

\bibitem{xian2018zero}
Y.~Xian, C.~H. Lampert, B.~Schiele, and Z.~Akata, ``Zero-shot learning—a
  comprehensive evaluation of the good, the bad and the ugly,'' \emph{IEEE
  transactions on pattern analysis and machine intelligence}, vol.~41, no.~9,
  pp. 2251--2265, 2018.

\bibitem{ramesh2021zero}
A.~Ramesh, M.~Pavlov, G.~Goh, S.~Gray, C.~Voss, A.~Radford, M.~Chen, and
  I.~Sutskever, ``Zero-shot text-to-image generation,'' in \emph{International
  Conference on Machine Learning}.\hskip 1em plus 0.5em minus 0.4em\relax PMLR,
  2021, pp. 8821--8831.

\bibitem{changpinyo2016synthesized}
S.~Changpinyo, W.-L. Chao, B.~Gong, and F.~Sha, ``Synthesized classifiers for
  zero-shot learning,'' in \emph{Proceedings of the IEEE conference on computer
  vision and pattern recognition}, 2016, pp. 5327--5336.

\bibitem{nagarajan2018attributes}
T.~Nagarajan and K.~Grauman, ``Attributes as operators: factorizing unseen
  attribute-object compositions,'' in \emph{Proceedings of the European
  Conference on Computer Vision (ECCV)}, 2018, pp. 169--185.

\bibitem{purushwalkam2019task}
S.~Purushwalkam, M.~Nickel, A.~Gupta, and M.~Ranzato, ``Task-driven modular
  networks for zero-shot compositional learning,'' in \emph{Proceedings of the
  IEEE/CVF International Conference on Computer Vision}, 2019, pp. 3593--3602.

\bibitem{misra2017red}
I.~Misra, A.~Gupta, and M.~Hebert, ``From red wine to red tomato: Composition
  with context,'' in \emph{Proceedings of the IEEE Conference on Computer
  Vision and Pattern Recognition}, 2017, pp. 1792--1801.

\bibitem{li2020symmetry}
Y.-L. Li, Y.~Xu, X.~Mao, and C.~Lu, ``Symmetry and group in attribute-object
  compositions,'' in \emph{Proceedings of the IEEE/CVF Conference on Computer
  Vision and Pattern Recognition}, 2020, pp. 11\,316--11\,325.

\bibitem{atzmon2020causal}
Y.~Atzmon, F.~Kreuk, U.~Shalit, and G.~Chechik, ``A causal view of
  compositional zero-shot recognition,'' \emph{Advances in Neural Information
  Processing Systems}, vol.~33, pp. 1462--1473, 2020.

\bibitem{Karthik_2022_CVPR}
S.~Karthik, M.~Mancini, and Z.~Akata, ``Kg-sp: Knowledge guided simple
  primitives for open world compositional zero-shot learning,'' in
  \emph{Proceedings of the IEEE/CVF Conference on Computer Vision and Pattern
  Recognition (CVPR)}, June 2022, pp. 9336--9345.

\bibitem{mancini2022learning}
M.~Mancini, M.~F. Naeem, Y.~Xian, and Z.~Akata, ``Learning graph embeddings for
  open world compositional zero-shot learning,'' \emph{IEEE Transactions on
  Pattern Analysis and Machine Intelligence}, 2022.

\bibitem{karthik2021revisiting}
S.~Karthik, M.~Mancini, and Z.~Akata, ``Revisiting visual product for
  compositional zero-shot learning,'' in \emph{NeurIPS 2021 Workshop on
  Distribution Shifts: Connecting Methods and Applications}, 2021.

\bibitem{huynh2020compositional}
D.~Huynh and E.~Elhamifar, ``Compositional zero-shot learning via fine-grained
  dense feature composition,'' \emph{Advances in Neural Information Processing
  Systems}, vol.~33, 2020.

\bibitem{yang2022decomposable}
M.~Yang, C.~Xu, A.~Wu, and C.~Deng, ``A decomposable causal view of
  compositional zero-shot learning,'' \emph{IEEE Transactions on Multimedia},
  2022.

\bibitem{li2022siamese}
X.~Li, X.~Yang, K.~Wei, C.~Deng, and M.~Yang, ``Siamese contrastive embedding
  network for compositional zero-shot learning,'' in \emph{Proceedings of the
  IEEE/CVF Conference on Computer Vision and Pattern Recognition}, 2022, pp.
  9326--9335.

\bibitem{zhang2022learning}
T.~Zhang, K.~Liang, R.~Du, X.~Sun, Z.~Ma, and J.~Guo, ``Learning invariant
  visual representations for compositional zero-shot learning,'' \emph{arXiv
  preprint arXiv:2206.00415}, 2022.

\bibitem{saini2022disentangling}
N.~Saini, K.~Pham, and A.~Shrivastava, ``Disentangling visual embeddings for
  attributes and objects,'' in \emph{Proceedings of the IEEE/CVF Conference on
  Computer Vision and Pattern Recognition}, 2022, pp. 13\,658--13\,667.

\bibitem{mancini2021open}
M.~Mancini, M.~F. Naeem, Y.~Xian, and Z.~Akata, ``Open world compositional
  zero-shot learning,'' in \emph{Proceedings of the IEEE/CVF conference on
  computer vision and pattern recognition}, 2021, pp. 5222--5230.

\bibitem{naeem20223d}
M.~F. Naeem, E.~P. {\"O}rnek, Y.~Xian, L.~Van~Gool, and F.~Tombari, ``3d
  compositional zero-shot learning with decompositional consensus,'' in
  \emph{European Conference on Computer Vision}.\hskip 1em plus 0.5em minus
  0.4em\relax Springer, 2022, pp. 713--730.

\bibitem{hou2020visual}
Z.~Hou, X.~Peng, Y.~Qiao, and D.~Tao, ``Visual compositional learning for
  human-object interaction detection,'' in \emph{European Conference on
  Computer Vision}.\hskip 1em plus 0.5em minus 0.4em\relax Springer, 2020, pp.
  584--600.

\bibitem{hou2021detecting}
Z.~Hou, B.~Yu, Y.~Qiao, X.~Peng, and D.~Tao, ``Detecting human-object
  interaction via fabricated compositional learning,'' in \emph{Proceedings of
  the IEEE/CVF Conference on Computer Vision and Pattern Recognition}, 2021,
  pp. 14\,646--14\,655.

\bibitem{kato2018compositional}
K.~Kato, Y.~Li, and A.~Gupta, ``Compositional learning for human object
  interaction,'' in \emph{Proceedings of the European Conference on Computer
  Vision (ECCV)}, 2018, pp. 234--251.

\bibitem{cho2017compositional}
N.-G. Cho, S.-H. Park, J.-S. Park, U.~Park, and S.-W. Lee, ``Compositional
  interaction descriptor for human interaction recognition,''
  \emph{Neurocomputing}, vol. 267, pp. 169--181, 2017.

\bibitem{materzynska2020something}
J.~Materzynska, T.~Xiao, R.~Herzig, H.~Xu, X.~Wang, and T.~Darrell,
  ``Something-else: Compositional action recognition with spatial-temporal
  interaction networks,'' in \emph{Proceedings of the IEEE/CVF Conference on
  Computer Vision and Pattern Recognition}, 2020, pp. 1049--1059.

\bibitem{wei2019adversarial}
K.~Wei, M.~Yang, H.~Wang, C.~Deng, and X.~Liu, ``Adversarial fine-grained
  composition learning for unseen attribute-object recognition,'' in
  \emph{Proceedings of the IEEE/CVF International Conference on Computer
  Vision}, 2019, pp. 3741--3749.

\bibitem{anwaar2021contrastive}
M.~U. Anwaar, R.~A. Khan, Z.~Pan, and M.~Kleinsteuber, ``A contrastive learning
  approach for compositional zero-shot learning,'' in \emph{Proceedings of the
  2021 International Conference on Multimodal Interaction}, 2021, pp. 34--42.

\bibitem{naeem2021learning}
M.~F. Naeem, Y.~Xian, F.~Tombari, and Z.~Akata, ``Learning graph embeddings for
  compositional zero-shot learning,'' in \emph{Proceedings of the IEEE/CVF
  Conference on Computer Vision and Pattern Recognition}, 2021, pp. 953--962.

\bibitem{lu2016visual}
C.~Lu, R.~Krishna, M.~Bernstein, and L.~Fei-Fei, ``Visual relationship
  detection with language priors,'' in \emph{European conference on computer
  vision}.\hskip 1em plus 0.5em minus 0.4em\relax Springer, 2016, pp. 852--869.

\bibitem{xu2021relation}
Z.~Xu, G.~Wang, Y.~Wong, and M.~S. Kankanhalli, ``Relation-aware compositional
  zero-shot learning for attribute-object pair recognition,'' \emph{IEEE
  Transactions on Multimedia}, 2021.

\bibitem{lian2022robust}
J.~Lian, C.~Zhang, and D.~Yu, ``Robust disentangled variational speech
  representation learning for zero-shot voice conversion,'' in \emph{ICASSP
  2022-2022 IEEE International Conference on Acoustics, Speech and Signal
  Processing (ICASSP)}.\hskip 1em plus 0.5em minus 0.4em\relax IEEE, 2022, pp.
  6572--6576.

\bibitem{jhoo2021collaborative}
W.~Y. Jhoo and J.-P. Heo, ``Collaborative learning with disentangled features
  for zero-shot domain adaptation,'' in \emph{Proceedings of the IEEE/CVF
  International Conference on Computer Vision}, 2021, pp. 8896--8905.

\bibitem{gabbay2021image}
A.~Gabbay, N.~Cohen, and Y.~Hoshen, ``An image is worth more than a thousand
  words: Towards disentanglement in the wild,'' \emph{Advances in Neural
  Information Processing Systems}, vol.~34, pp. 9216--9228, 2021.

\bibitem{li2022zero}
J.~Li, Z.~Ling, L.~Niu, and L.~Zhang, ``Zero-shot sketch-based image retrieval
  with structure-aware asymmetric disentanglement,'' \emph{Computer Vision and
  Image Understanding}, vol. 218, p. 103412, 2022.

\bibitem{tong2019hierarchical}
B.~Tong, C.~Wang, M.~Klinkigt, Y.~Kobayashi, and Y.~Nonaka, ``Hierarchical
  disentanglement of discriminative latent features for zero-shot learning,''
  in \emph{Proceedings of the IEEE/CVF Conference on Computer Vision and
  Pattern Recognition}, 2019, pp. 11\,467--11\,476.

\bibitem{fan2022contrastive}
W.~Fan, C.~Liang, and T.~Wang, ``Contrastive semantic disentanglement in latent
  space for generalized zero-shot learning,'' \emph{Knowledge-Based Systems},
  p. 109949, 2022.

\bibitem{chen2021semantics}
Z.~Chen, Y.~Luo, R.~Qiu, S.~Wang, Z.~Huang, J.~Li, and Z.~Zhang, ``Semantics
  disentangling for generalized zero-shot learning,'' in \emph{Proceedings of
  the IEEE/CVF international conference on computer vision}, 2021, pp.
  8712--8720.

\bibitem{li2021generalized}
X.~Li, Z.~Xu, K.~Wei, and C.~Deng, ``Generalized zero-shot learning via
  disentangled representation,'' in \emph{Proceedings of the AAAI Conference on
  Artificial Intelligence}, vol.~35, no.~3, 2021, pp. 1966--1974.

\bibitem{li2021disentangled}
B.~Li, C.~Han, T.~Guo, and T.~Zhao, ``Disentangled features with direct sum
  decomposition for zero shot learning,'' \emph{Neurocomputing}, vol. 426, pp.
  216--226, 2021.

\bibitem{geng2022disentangled}
Y.~Geng, J.~Chen, W.~Zhang, Y.~Xu, Z.~Chen, J.~Z.~Pan, Y.~Huang, F.~Xiong, and
  H.~Chen, ``Disentangled ontology embedding for zero-shot learning,'' in
  \emph{Proceedings of the 28th ACM SIGKDD Conference on Knowledge Discovery
  and Data Mining}, 2022, pp. 443--453.

\bibitem{yu2014fine}
A.~Yu and K.~Grauman, ``Fine-grained visual comparisons with local learning,''
  in \emph{Proceedings of the IEEE conference on computer vision and pattern
  recognition}, 2014, pp. 192--199.

\bibitem{yu2017semantic}
------, ``Semantic jitter: Dense supervision for visual comparisons via
  synthetic images,'' in \emph{Proceedings of the IEEE International Conference
  on Computer Vision}, 2017, pp. 5570--5579.

\bibitem{isola2015discovering}
P.~Isola, J.~J. Lim, and E.~H. Adelson, ``Discovering states and
  transformations in image collections,'' in \emph{Proceedings of the IEEE
  conference on computer vision and pattern recognition}, 2015, pp. 1383--1391.

\bibitem{hudson2019gqa}
D.~A. Hudson and C.~D. Manning, ``Gqa: A new dataset for real-world visual
  reasoning and compositional question answering,'' in \emph{Proceedings of the
  IEEE/CVF conference on computer vision and pattern recognition}, 2019, pp.
  6700--6709.

\bibitem{xu2020attribute}
W.~Xu, Y.~Xian, J.~Wang, B.~Schiele, and Z.~Akata, ``Attribute prototype
  network for zero-shot learning,'' in \emph{34th Conference on Neural
  Information Processing Systems}.\hskip 1em plus 0.5em minus 0.4em\relax
  Curran Associates, Inc., 2020.

\bibitem{xie2019attentive}
G.-S. Xie, L.~Liu, X.~Jin, F.~Zhu, Z.~Zhang, J.~Qin, Y.~Yao, and L.~Shao,
  ``Attentive region embedding network for zero-shot learning,'' in
  \emph{Proceedings of the IEEE Conference on Computer Vision and Pattern
  Recognition}, 2019, pp. 9384--9393.

\bibitem{chao2016empirical}
W.-L. Chao, S.~Changpinyo, B.~Gong, and F.~Sha, ``An empirical study and
  analysis of generalized zero-shot learning for object recognition in the
  wild,'' in \emph{European conference on computer vision}.\hskip 1em plus
  0.5em minus 0.4em\relax Springer, 2016, pp. 52--68.

\bibitem{mikolov2013distributed}
T.~Mikolov, I.~Sutskever, K.~Chen, G.~S. Corrado, and J.~Dean, ``Distributed
  representations of words and phrases and their compositionality,''
  \emph{Advances in neural information processing systems}, vol.~26, 2013.

\bibitem{bojanowski2017enriching}
P.~Bojanowski, E.~Grave, A.~Joulin, and T.~Mikolov, ``Enriching word vectors
  with subword information,'' \emph{Transactions of the association for
  computational linguistics}, vol.~5, pp. 135--146, 2017.

\bibitem{he2016deep}
K.~He, X.~Zhang, S.~Ren, and J.~Sun, ``Deep residual learning for image
  recognition,'' in \emph{Proceedings of the IEEE conference on computer vision
  and pattern recognition}, 2016, pp. 770--778.

\bibitem{kingma2014adam}
D.~P. Kingma and J.~Ba, ``Adam: A method for stochastic optimization,''
  \emph{arXiv preprint arXiv:1412.6980}, 2014.

\bibitem{krogh1991simple}
A.~Krogh and J.~Hertz, ``A simple weight decay can improve generalization,''
  \emph{Advances in neural information processing systems}, vol.~4, 1991.

\bibitem{paszke2019pytorch}
A.~Paszke, S.~Gross, F.~Massa, A.~Lerer, J.~Bradbury, G.~Chanan, T.~Killeen,
  Z.~Lin, N.~Gimelshein, L.~Antiga \emph{et~al.}, ``Pytorch: An imperative
  style, high-performance deep learning library,'' \emph{Advances in neural
  information processing systems}, vol.~32, 2019.

\bibitem{cuda}
J.~Nickolls, I.~Buck, M.~Garland, and K.~Skadron, ``Scalable parallel
  programming with cuda: Is cuda the parallel programming model that
  application developers have been waiting for?'' \emph{Queue}, vol.~6, no.~2,
  pp. 40--53, 2008.

\bibitem{van2008visualizing}
L.~Van~der Maaten and G.~Hinton, ``Visualizing data using t-sne.''
  \emph{Journal of machine learning research}, vol.~9, no.~11, 2008.

\bibitem{jin2021channel}
Y.~Jin, J.~Zhang, X.~Zhang, H.~Xiao, B.~Ai, and D.~W.~K. Ng, ``Channel
  estimation for semi-passive reconfigurable intelligent surfaces with enhanced
  deep residual networks,'' \emph{IEEE transactions on vehicular technology},
  vol.~70, no.~10, pp. 11\,083--11\,088, 2021.

\bibitem{zhang2022deep}
B.~Zhang and C.~Gao, ``Deep residual network for image super-resolution
  reconstruction,'' in \emph{2022 12th International Conference on CYBER
  Technology in Automation, Control, and Intelligent Systems (CYBER)}.\hskip
  1em plus 0.5em minus 0.4em\relax IEEE, 2022, pp. 620--623.

\end{thebibliography}

\end{document}